\def\year{2022}\relax
\documentclass[letterpaper]{article} % DO NOT CHANGE THIS
\usepackage{aaai22}  % DO NOT CHANGE THIS
\usepackage{times}  % DO NOT CHANGE THIS
\usepackage{helvet}  % DO NOT CHANGE THIS
\usepackage{courier}  % DO NOT CHANGE THIS
\usepackage[hyphens]{url}  % DO NOT CHANGE THIS
\usepackage{graphicx} % DO NOT CHANGE THIS
\usepackage{natbib}  % DO NOT CHANGE THIS AND DO NOT ADD ANY OPTIONS TO IT
\usepackage{caption} % DO NOT CHANGE THIS AND DO NOT ADD ANY OPTIONS TO IT
\DeclareCaptionStyle{ruled}{labelfont=normalfont,labelsep=colon,strut=off} % DO NOT CHANGE THIS
\usepackage{algorithm}
\usepackage{algorithmic}
\usepackage{color}

\newcommand{\LJW}[1]{\textcolor{green}{{#1}}}

\usepackage{newfloat}
\usepackage{appendix}

\usepackage{listings}
\usepackage{multirow}
\floatstyle{ruled}
\newfloat{listing}{tb}{lst}{}
\floatname{listing}{Listing}
\usepackage{amsfonts}

\title{Pose Guided Image Generation from Misaligned Sources via Residual Flow Based Correction}

\author{
    Jiawei Lu\textsuperscript{\rm 1}, He Wang\textsuperscript{\rm 2}, Tianjia Shao\textsuperscript{\rm 1}\thanks{Corresponding author.}, Yin Yang\textsuperscript{\rm 3}, Kun Zhou\textsuperscript{\rm 1}
}
\affiliations{
    \textsuperscript{\rm 1} State Key Lab of CAD\&CG, Zhejiang University\\
    \textsuperscript{\rm 2} University of Leeds\\
    \textsuperscript{\rm 3} Clemson University\\
    lujiawei23@gmail.com,
    h.e.wang@leeds.ac.uk,
    tjshao@zju.edu.cn,
    yin5@clemson.edu,
    kunzhou@zju.edu.cn
}
\let\oldtwocolumn\twocolumn
\renewcommand\twocolumn[1][]{%
    \oldtwocolumn[{#1}{
    \begin{center}
           \includegraphics[width=\textwidth]{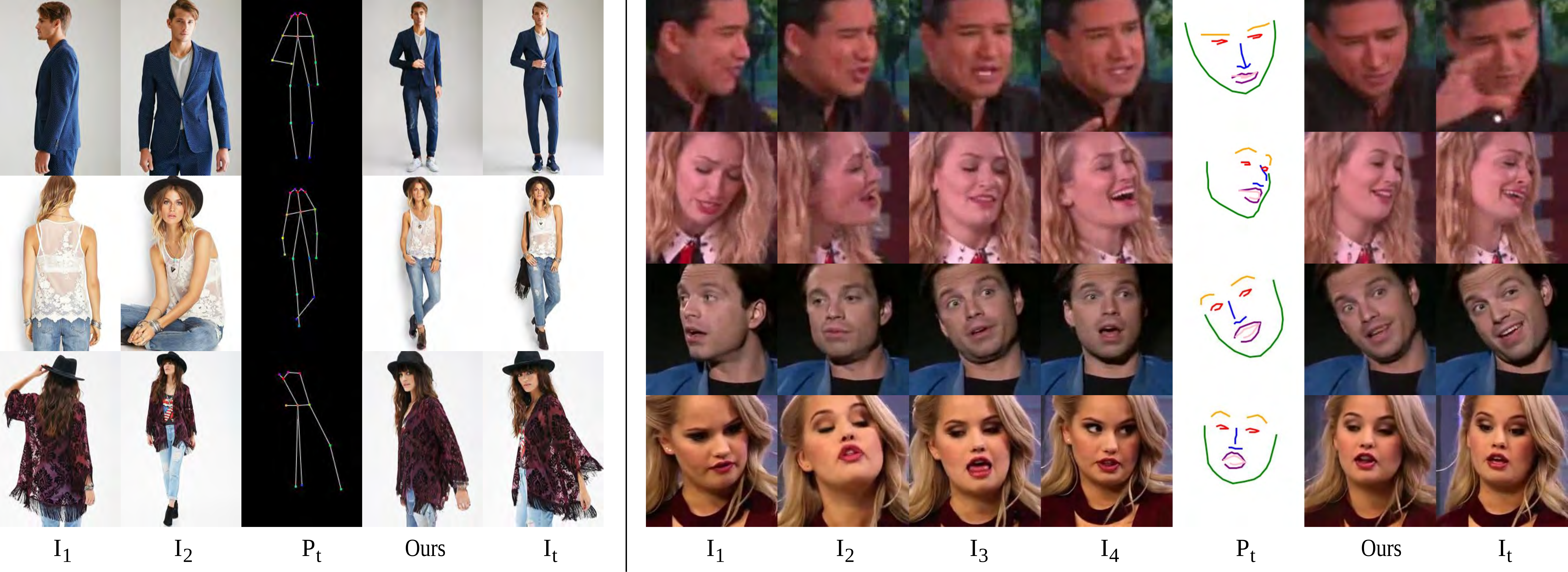}
           \captionof{figure}{Image generation on body poses (left) and facial expressions (right). Our model takes an arbitrary number of misaligned source images (I$_1$-I$_4$) and a target pose (P$_t$) to generate new images (Ours). I$_t$ is the target ground truth.}
           \label{fig:teaser}
        \end{center}
    }]
}
\begin{document}
\maketitle
% \twocolumn[{%
% \renewcommand\twocolumn[1][]{#1}%
% \maketitle
% \begin{center}
%     \centering
%     \captionsetup{type=figure}
%     \includegraphics[width=.95\textwidth]{teaser}
%     \captionof{figure}{Image generation on body poses (left) and facial expressions (right). Our model takes an arbitrary number of misaligned source images (I$_1$-I$_4$) and a target pose (P$_t$) to generate new images (Ours). I$_t$ is the target ground truth.}
%     \label{fig:teaser}
% \end{center}%
% }]

% \begin{strip}
% \centering
% \includegraphics[width=0.9\textwidth]{teaser.pdf}
% \captionsetup{type=figure}
% \captionof{figure}{Image generation on body poses (left) and facial expressions (right). Our model takes an arbitrary number of misaligned source images (I$_1$-I$_4$) and a target pose (P$_t$) to generate new images (Ours). I$_t$ is the target ground truth.}
% \label{fig:teaser}
% \end{strip}

% \begin{figure*}
%   \includegraphics[width=1.0\linewidth]{teaser}
% \end{figure*}

\begin{abstract}
Generating new images with desired properties (e.g. new view/poses) from source images has been enthusiastically pursued recently, due to its wide range of potential applications. One way to ensure high-quality generation is to use multiple sources with complementary information such as different views of the same object. However, as source images are often misaligned due to the large disparities among the camera settings, strong assumptions have been made in the past with respect to the camera(s) or/and the object in interest, limiting the application of such techniques. Therefore, we propose a new general approach which models multiple types of variations among sources, such as view angles, poses, facial expressions, in a unified framework, so that it can be employed on datasets of vastly different nature. We verify our approach on a variety of data including humans bodies, faces, city scenes and 3D objects. Both the qualitative and quantitative results demonstrate the better performance of our method than the state of the art. 
\end{abstract}

\section{Introduction}
% \noindent Pose-guided image generation has aroused extensive attention due to its wide range of applications such as person image generation~\cite{??}, facial image generation~\cite{??} and scene image generation\cite{??}.  A number of works~\cite{NIPS2017_34ed066d, Siarohin_2018_CVPR, zhou2016view} have achieved great success on single-source posed-guided image generation, which synthesizes a new image with a different pose/view from a single source image. However, single-source image generation is inherently an ill-posed problem as a lot information is missing due to large pose/view differences and self-occlusion (e.g., generating the back view of a person given the front view). As a result, it is still challenging to generate a high-quality image from a single source.

\noindent Controlled image generation from source images is capable of generating scenes in unseen views and objects with pre-defined poses. It has a wide range of applications, e.g people with new poses~\cite{NIPS2017_34ed066d}, faces with new expressions~\cite{Zakharov_2019_ICCV} and scenes from different angles~\cite{sun2018multiview}, and hence has attracted attention. The key challenge in such research is to recover the hidden information from sparse view points. One popular setting is to employ a single source image to generate new images with new poses/views. Despite recent successes~\cite{NIPS2017_34ed066d, Siarohin_2018_CVPR, zhang2021pise}, ambiguity caused by the limited information available in a single image still makes it difficult to synthesize a high-quality image with large pose differences (e.g. generating the back view of a person when given only the front view). Consequently, high-quality generation is still an open challenge.

% To tackle this challenge, a natural way is to use multiple source images as input, as multiple sources with different poses/views can offer information complementary to each other. A variety of works have been done in this area. For example, \cite{sun2018multiview} and \cite{ zhou2016view} take multi-view images of a rigid object or scene to synthesize its novel view. However, they cannot handle objects with pose variances (e.g. human body). Zakharov et al.~\cite{Zakharov_2019_ICCV} utilize multiple face images to generate a new face with desired expression, but they encode multiple faces into a single style vector, which may lose the details of source images. Lathuilière et al.~\cite{lathuiliere2020attention} make use of several human images to produce a new human image with a different pose, which \tianjia{applies affine transformation to warp the source features into target pose}. However, they may fail to generate vivid details as affine transformation can be insufficient for large pose variations, especially for the large non-rigid motions of clothes. 

In theory, employing multiple source images with complementary information should mitigate the problem. However, in practice, this setting unfortunately brings additional challenges: the source images especially in-the-wild ones are not taken by calibrated cameras, leading to severe misalignment. Given the huge size of the camera space (possible camera poses), it is not straightforward to design a general solution. Consequently, strong assumptions have to be made. If the object is assumed to be rigid, multi-view images can help synthesize novel views~\cite{sun2018multiview,zhou2016view}. If deformable objects are involved, certain types of alignment or transformations need to be assumed, such as feature averaging can help synthesize new facial expressions~\cite{Zakharov_2019_ICCV}, but at the cost of losing the details of the source images; affine transformation can help synthesize new human poses~\cite{lathuiliere2020attention}, but incapable of handling large pose deformation especially for non-rigid deformation such as clothes.

In this paper, we seek a general framework for controlled multi-source image generation. The framework takes as input multiple (misaligned) source images and source poses as well as a target pose, and predicts a new image under the target pose while keeping the source appearance. One key challenge is to impose parsimonious assumptions on the source images, so that they can differ in the camera pose, the camera-object distance, occlusions/lighting, etc. We aim to simultaneously deal with view/pose/expression variances and meanwhile synthesize high quality images with realistic details. One intuitive solution is to warp the features of each source image then fuse them for the target pose. However, two issues appear in such an approach. First, each source image is only a partial observation of the object, and all source images are misaligned. As a result, fusing such features inevitably leads to blurring in the target image. Second, the partial observability essentially dictates that different regions in a source image provides information with different levels of confidence (i.e. occluded areas having low confidence). Further, source images have different importance, so do their high/low confidence areas by association. This hierarchical structure of importance among source images cannot be captured via simple treatments, e.g. an occlusion map on a source image~\cite{ren2020deep}, or attention maps merely distinguishing the relative importance of different views~\cite{sun2018multiview}.

To tackle the above challenges, we propose a novel fusion mechanism. Our framework adopts the state-of-the-art flow-based strategy, which first learns to warp each source feature to match the target pose at different levels, and then fuses these features in a decoder to synthesize the image. To tackle the challenge of hierarchical feature confidence, we propose to simultaneously predict the attention map and occlusion map for each source in the source feature extractor. The attention maps indicate the important source regions, and the occlusion maps dictate which part is invisible and should be inpainted. This way, the warped source features can be fused while being aware of the confidences of different source parts and invisible regions. To address the feature misalignment issue, we propose a novel residual-fusing (RF) block to correct the warping. A RF block consists of two modules: residual module and fusing module. The residual module corrects the warping of the source features and learns a residual flow for each warped source feature to match the fused feature from previous level. The fuse module takes the output of the residual module, and corrects the warped feature via an occlusion map. Then the corrected features are further fused by attention maps and sent to the next block. Overall, the RF blocks are repeated many times at multiple feature layers, so that different source features can be warped into a consistent space and be decoded to generate images with less artifacts and blurring.

Formally, our contributions include:
\begin{itemize}
\item a new general framework for controlled multi-source image generation, which can effectively capture view/pose/expression variances and synthesize high quality images with realistic details. 

\item a new Residual-Fusing block to systematically reconcile the conflicts caused by the misalignment of multiple (calibrated) sources.

\item comprehensive experiments and comparisons on multiple distinctive datasets across different tasks to demonstrate the superiority of multi-source image generation under our general framework.
\end{itemize}

\begin{figure*}[t]
\centering
\includegraphics[width=1.0\linewidth]{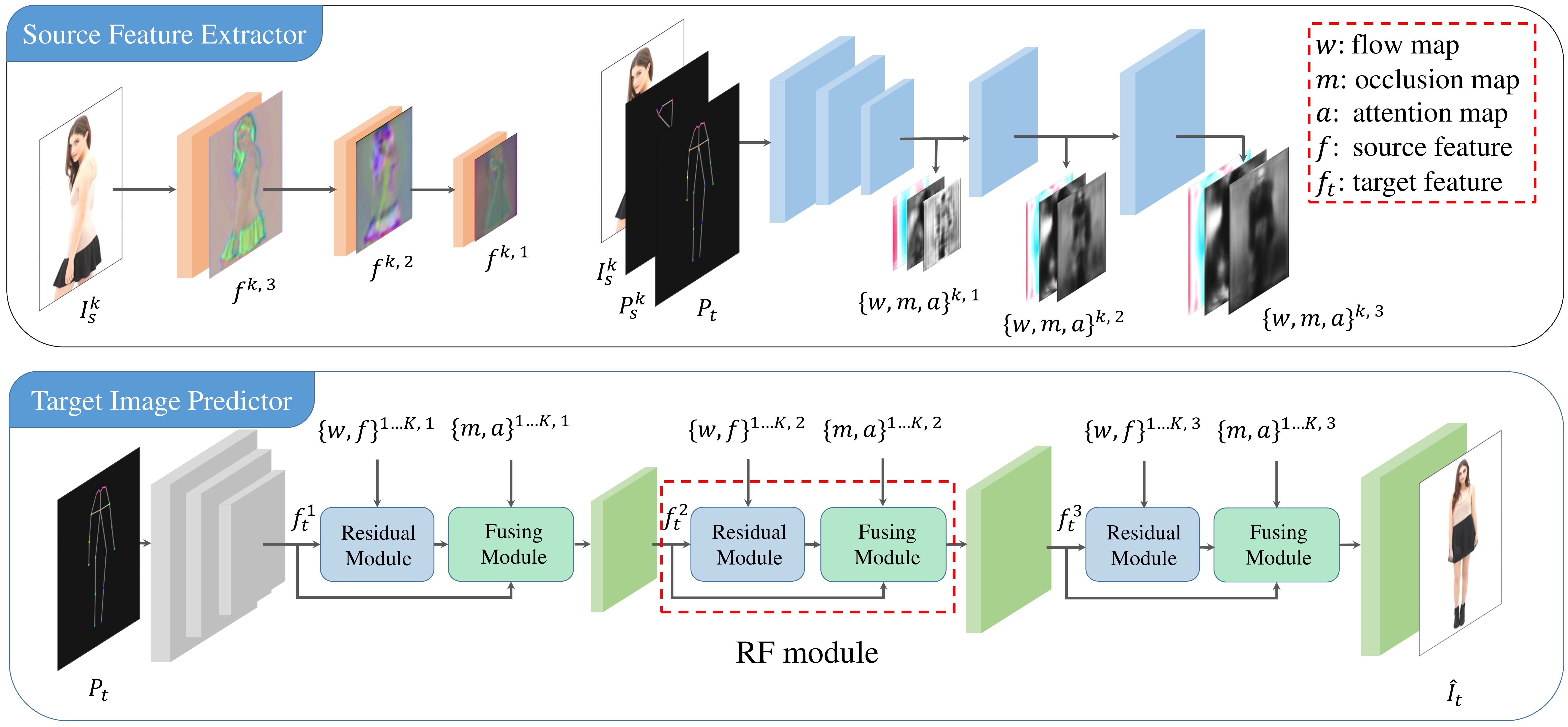}
\caption{Network overview. Given $K$ input images \{$I_s^k$\}, their poses \{$P_s^k$\} and target pose $P_t$, we aim to generate a new image $\hat{I}_t$ in target pose $P_t$. Our source feature extractor first extracts the image features \{$f^{k}$\} and generates flow maps \{$w^k$\}, occlusion maps \{$m^k$\}, attention maps \{$a^k$\} at different levels for each source independently. The Target Image Predictor first encodes the target pose into target feature $f_t^1$. Then, the proposed Residual-Fusing Module in the decoder part of Target Image Predictor repeats at different feature levels to gradually fuse the sources and generate the target image $\hat{I}_t$.}
\label{fig_pipline}
\end{figure*}

\section{Related work}
\subsubsection{Single Source Image Generation.}
Single source image generation aims to synthesize new images given a source image and a target pose. It involves many tasks including human pose transfer, novel view synthesis, facial image generation, etc. Ma et al.~\shortcite{NIPS2017_34ed066d} first introduced pose-guided human image generation and proposed a two-stage adversarial framework, which first synthesizes a coarse person image and then refines the result. Balakrishnan et al.~\shortcite{balakrishnan2018synthesizing} presented a modular GAN network which decouples different body parts into layers and moves them by affine transformation. Siarohin et al.~\shortcite{Siarohin_2018_CVPR} applied affine transformations in feature space by deformable skip connections. Zhu et al.~\shortcite{zhu2019progressive} proposed a novel block which simultaneously updates the pose code and appearance code in a coarse to fine manner. Although these works can synthesize correct global structures, they fail to preserve the local texture details provided in source images. In contrast, flow-based methods can better transfer the details such as clothing and texture. Han et al.~\shortcite{Han_2019_ICCV} proposed a three-stage network which first generates a semantic parsing map and then learns a flow of each semantic region. However, an extra refinement network is required as they predict the flow at the pixel level. Ren et al.~\shortcite{ren2020deep} presented a global flow and local attention architecture to generate vivid textures, but they struggled to synthesize unseen regions from a single source image. 
% \textbf{Warping-based Image Synthesis.} \quad
% Our approach is closely related to warping-based image generation method.\\
\subsubsection{Multi-source Image Generation.}
Our work is closely related to multi-source image generation methods. Zhou et al.~\shortcite{zhou2016view} utilized multiple views of an object to generate novel view given a target camera pose. They predict a pixel flow map together with a confidence map for each single view and merge them together by confidence maps. Sun et al.~\shortcite{sun2018multiview} improved ~\cite{zhou2016view} by adding a convolutional LSTM generator~\cite{xingjian2015convolutional} to hallucinate the missing pixels from source view. Inspired by them, we also employ confidence maps and target generator in our work, but there are two main differences between our work and ~\cite{sun2018multiview}. First, Sun et al.~\shortcite{sun2018multiview} conducted experiment mainly on rigid objects such as cars and chairs, while we can handle more general tasks, including highly non-rigid human image generation and facial image transfer. Second, they aggregated source images at pixel level directly, while we found it inappropriate in more general cases due to the complex texture and large motion, thus we propose to warp multi-level features instead of warping pixels. Lathuiliere et al~\shortcite{lathuiliere2020attention} introduced an attention-based decoder for multi-source human pose transfer. Some other works~\cite{Chan_2019_ICCV, wang2018vid2vid, liu2019neural} had to train a model for each target, which limits their applications. Wang et al.~\shortcite{wang2018fewshotvid2vid} improved \cite{wang2018vid2vid} by generating network weights dynamically from reference images, however, continuous videos are required to calculate optical flow. In contrast, our approach is more general and can deal with flexible number of inputs with arbitrary poses.

\section{Method}
\subsection{Overview}
We propose a general generative adversarial network for multi-source pose guided image generation. Here the `pose' can be any structural information of an image, e.g. human joints, view angles, facial landmarks, etc. %To the best of our knowledge, no prior work can handle this task under such a general setting. Sun et al.~\shortcite{sun2018multiview} can generate rigid objects and scenes with the guidance of a new view and multiple sources, but fail in objects are highly deformable such as human poses. Zakharov et al.~\shortcite{Zakharov_2019_ICCV} focus on realistic talking faces synthesis \HW{but?}. Lathuiliere et al.~\shortcite{lathuiliere2020attention} dedicate on multi-source person image generation. \HW{but?} 
Let $K$ be the number of sources. Our generator $G$ takes $\{I_s^k, P_s^k\}_{k=1,...,K}$ and $P_t$ as input, where $I_s^k$ denotes the $i$-th input image, $P_s^k$ denotes the corresponding pose representation, and $P_t$ denotes the target pose. Our goal is to synthesize a new image $\hat{I}_t$ matching the target pose and meanwhile keep the source appearance. $G$ can be written as:
\begin{equation}
    \hat{I}_t=G(\{I_s^k\},\{P_s^k\}, P_t).
\end{equation}

Now we give a general overview of our architecture. Our model consists of two parts: source feature extractor and target image predictor. As shown in Fig.~\ref{fig_pipline}, in the source feature extractor, for each source, we estimate initial flow maps for warping the source features at different levels, and simultaneously predict the corresponding attention maps and occlusion maps. With the initial flow maps, the warped source features can be roughly aligned with the target pose at different levels. The necessity of multi-level modeling is primarily because misaligned features exist globally e.g. human poses, as well as locally e.g. textures on clothing. During the fusion, the attention maps play a role to select the important source regions among different sources, and the occlusion maps indicate which part is invisible and should be inpainted. As these source features are warped to match the target pose which only provides sparse structural information, directly fusing them will inevitably cause feature misalignment, leading to artifacts such as blurring and ghosting. Therefore, in the target image predictor, the residual-fusing (RF) module is brought up to further correct the warped features and fuse them. It contains two sub-modules: residual module and fusing module. At each feature level, the residual module takes the initially warped feature from each source branch, and learns a residual flow to further warp the feature to match the target feature from the previous level. The corrected source features are further sent to the fusing module, which performs a weighted aggregation of different sources using the occlusion maps and attention maps, and outputs the fused target feature to the next level. 

%We now give a detailed discussion on each part of our model.

\subsection{Source Feature Extractor}
The source feature extractor $F$ takes ${I_s^k, P_s^k, P_t}$ as input and generates the initial flow field $w^{k}$, attention map $a^{k}$ and occlusion map $m^{k}$ (as shown on the top right of Fig.~\ref{fig_pipline}):
\begin{equation}
    w^{k}, a^{k}, m^{k} = F({I_s^k, P_s^k, P_t}),
\end{equation}
where $w^{k}$ stores the coordinate displacements between the source and the target features, and $a^{k}$ and $m^{k}$ has continuous values between 0 and 1. $m^k$ measures how the target feature is visible in a source at a certain position, and $a^k$ indicates which source is more relevant to the target at a certain position. We design $F$ as a fully convolutional network with a pyramid architecture, which outputs $w^{k}, a^{k}$ and $m^{k}$ at $N$ different resolutions, i.e. $w^{k}, a^{k}, m^{k} = \{w^{k,i}, a^{k,i}, m^{k,i}\}, i=1...N$. $w^{k,i}, a^{k,i}$ and $m^{k,i}$ share the same backbone of $F$ except their output layers. Source image feature $f^{k}$ is extracted by another convolutional network(shown on the top left of Fig.~\ref{fig_pipline}). Please refer to the supplementary material for details. The attention map and occlusion map will be jointly applied in the subsequent Fusing Module to ensure a globally consistent feature fusion. 

% The sampling correctness loss is defined as follow:
% \begin{equation}
%     \mathcal{L}_{samp} = \frac{1}{N}\sum_{l\in \Omega }{exp(-\frac{\mu(v_{s,w}^l, v_t^l)}{\mu_{max}^l})}
% \end{equation}
% where $\mu({*})$ denotes the cosine similarity. $v_{s,w}^t$ and $v_t^l$ denotes VGG features of warped source image and target image at coordinates l, respectively. $\mu_{max}^l$ is a normalization term which calculates the max cosine similarity at target position $l$.

% With the flow field $w^{k,i}$ estimated for each source, we can warp the source features to the target pose by backward warping using bi-linear sampling:

% \begin{equation}
%     f_{w}^{k,i} = \mathcal{W}(f^{k,i}, w^{k,i}),
% \end{equation}
% where $\mathcal{W}$ denotes the backward warping operation, and $f^{k,i}$ denotes the extracted source image feature (as shown on the top left of Fig.~\ref{fig_pipline}). 

\subsection{Target Image Predictor}
After feature extraction, the image prediction is handled in the target image predictor. The major difficulty here is to fuse the source features into one consistent target feature and meanwhile reduce the feature misalignment from the initial warping. As shown on the bottom of Fig.~\ref{fig_pipline}, the target image generator starts from the target pose $P_t$ and then goes through several down-sampling layers to get the initial target feature $f_t^{1}$. Then, at each feature level $i$ in the predictor, a Residual-Fusing (RF) block is deployed to correct the warping field of each source feature based on the target feature from the previous level, and then fuse different sources together to output the target feature to the next level $i+1$.

\begin{figure}[t]
\centering
\includegraphics[width=1.0\linewidth]{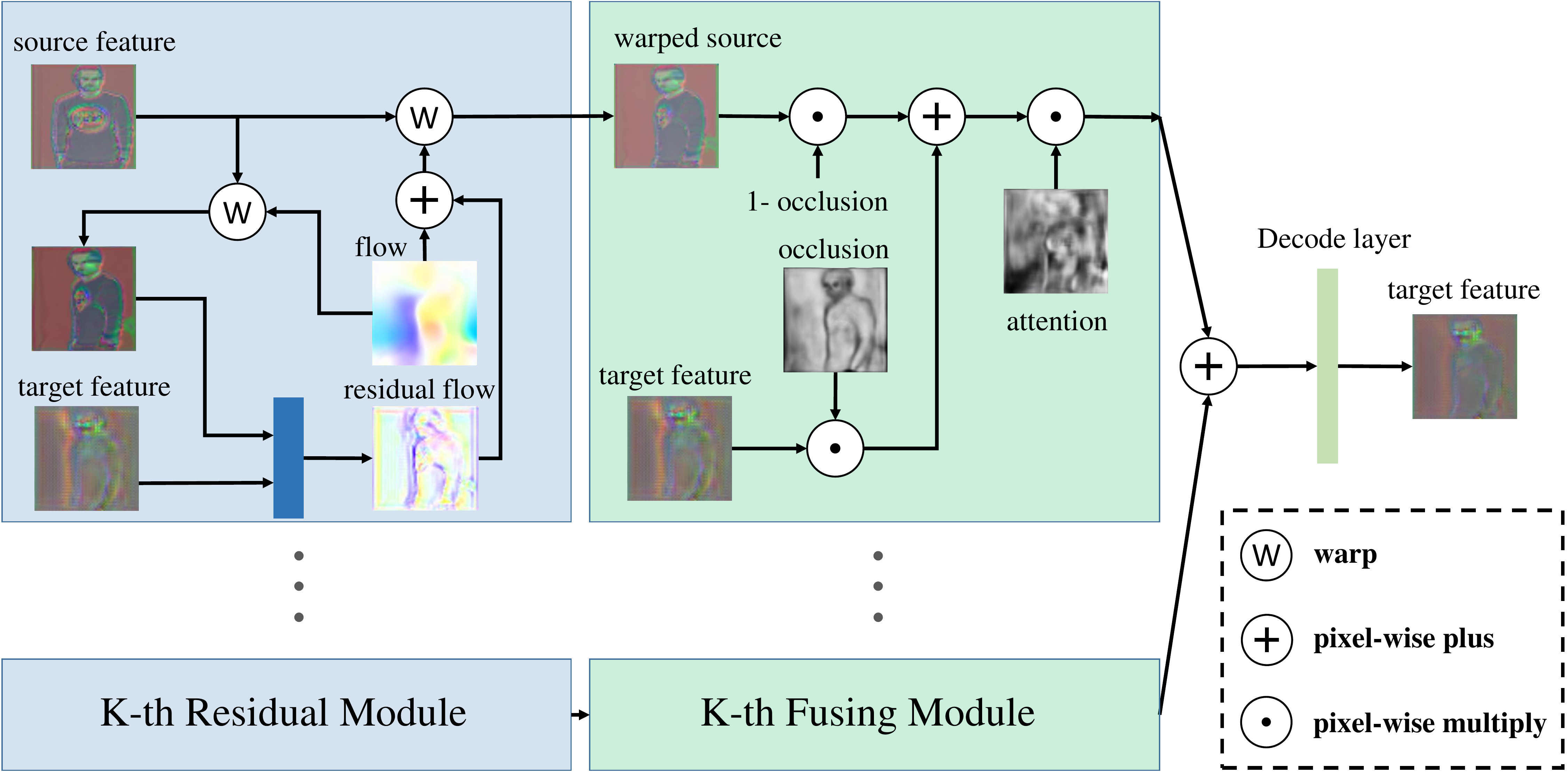}
\caption{The residual-fusing(RF) module repeats at different feature levels and has $K$ same branches. For a specific branch at level $i$, the residual module computes the residual flow to further warp the source feature. Then, the fusing module performs feature matting between the warped source feature and previous target feature by the occlusion map. Finally, $K$ features are fused together by the attention maps and output to the next level. Zoom in for better details.}
% \caption{The residual-fusing(RF) module repeats at different feature levels and has $K$ branches to deal with $K$ different sources. At each level $i$, the RF module receives source feature $f^{k,i}$, flow map $w^{k,i}$, attention map $a^{k,i}$ and occlusion map $m^{k,i}$ from the $K$-th source feature extractor, together with the previous fused feature $f_t^i$, to produce the next level fused feature $f_t^{i+1}$. Specifically, the $k$-th residual module first predict a residual flow based on the warped $k$-th source feature and previous fused feature, and output a refined source feature. Then, the fusing module merge the refined source feature and the previous fusing feature based on the occlusion map $m^{k,i}$. Finally, $K$ merged features are fused by $K$ attention maps and send to the decoder $D^i$ to produce the next level fused feature $f_t^{i+1}$.}
\label{fig_decoder}
\end{figure}

The RF block repeats at different feature levels and has $K$ branches to deal with $K$ different sources. It can be further divided into two sub-modules, named residual module and fusing module (as shown in Fig.~\ref{fig_decoder}). Now we take feature level $i$ as example and give a detailed description of each sub-module.

\subsubsection{Residual Module.}
The source feature extractor provides a coarse flow field to warp the source to match the target pose, which only provides sparse structural information. Directly fusing the coarsely warped features inevitable causes misalignment and artifacts. The residual module gives the network the ability to further correct the initial flow, by learning a residual flow from the initially warped source feature and the fused feature from previous level. 

At feature level $i$, the $k$-th residual module receives the source image feature $f^{k,i}$, and the flow map $w^{k,i}$ from the $k$-th source feature extractor, together with the fused feature $f_t^i$ from previous level. We first warp the source feature $f^{k,i}$ with the initial flow map $w^{k,i}$ to the target pose by:
\begin{equation}
    f_{w}^{k,i} = \mathcal{W}(f^{k,i}, w^{k,i}).
\end{equation}
Then we predict a residual flow $r^{k,i}$ by a residual flow predictor $R_i$ from $f_t^i$ and $f_w^{k,i}$:
\begin{equation}
    r^{k,i} = R^{i}(f_{t}^{i}, f_w^{k,i}).
\end{equation}
Note that $R^i$ is designed to share weights across different branches, making it possible for the model to accept arbitrary number of inputs at inference time.

With the learned residual flow, we get the refined flow field by adding the residual flow to the initial flow. We then warp each source feature $f^{k,i}$ by the refined flow and get the refined warped feature $\hat{f}_w^{k,i}$ of source branch $k$:
\begin{equation}
    \hat{f}_w^{k,i} = \mathcal{W}(f^{k,i}, w^{k,i} + r^{k,i}).
\end{equation}

\subsubsection{Fusing Module.}
With the refined warped source feature $\hat{f}_w^{k,i}$, the fusing module merges $\hat{f}_w^{k,i}$ and the previous fused target feature ${f}_t^{i}$ using the occlusion map $m^{k,i}$. Then, these merged features of $K$ branches are fused together into one, by a weighted summation over the $K$ attention maps $\{a^{k,i}\}$, where $a^{k,i}$ are normalized by a softmax operation at pixel level to stabilize the gradient during training. Finally, the fused feature goes through a decode layer $D^i$ to output the next level target feature $f_t^{i+1}$:
\begin{equation}
    f_{t}^{i+1} = D^{i}(\sum_{i=1}^{K}{a^{k,i} \cdot (\hat{f}_w^{k,i} \cdot (1-m^{k,i}) + f_{t}^{i} \cdot m^{k,i} ))}
\end{equation}

\subsection{Training}
We train our model in two stages. First, without the ground truth flow field, we warm up the flow generator $F$ using the sample correctness loss \cite{ren2019structureflow}. We also take the regularization loss \cite{ren2020deep} to constrain the smoothness of the flow.
\begin{equation}
    \mathcal{L}_{flow} = \lambda_{cor}\mathcal{L}_{cor} + \lambda_{reg}\mathcal{L}_{reg}
\end{equation}
where the sampling correctness loss $\mathcal{L}_{cor}$ maximizes the cosine similarity between the VGG features of the warped source and target and force the flow field $w^{k,i}$ to sample the similar regions. The regularization term $\mathcal{L}_{reg}$ penalize the local regions where the transformation is not an affine transformation. Then, with the pre-trained flow generator, we train our full model in an end-to-end manner.
The full loss can be defined as:
\begin{equation}
    \mathcal{L} = \mathcal{L}_{flow} + \mathcal{L}_{con} + \mathcal{L}_{adv} 
\end{equation}
where $\mathcal{L}_{con}$ is a content loss, and $\mathcal{L}_{con} = \lambda_{l_1}\mathcal{L}_{l_1} + \lambda_{per}\mathcal{L}_{per} + \lambda_{sty}\mathcal{L}_{sty}$. $\mathcal{L}_{l_1}$ minimizes the $L_1$ distance of the generated image and target image, $\mathcal{L}_{l_1} = ||\hat{I}_t - I_t  ||_1$. $\mathcal{L}_{per}$ and $\mathcal{L}_{sty}$ are inspired by~\cite{Johnson2016Perceptual}. The perceptual loss $\mathcal{L}_{per}$ aims to penalize the $L_1$ distance between features extracted from specific layers of a pre-traind VGG network: 
\begin{equation}
    \mathcal{L}_{per} = \sum_i||\phi_{i}(\hat{I}_t) - \phi_{i}(I_t)  ||_1
\end{equation}
where $\phi_{i}$ denotes the i-th layer of the VGG-19 network. The style loss $\mathcal{L}_{sty}$ uses the Gram matrix of VGG features to maximize the style similarity between the images:
\begin{equation}
    \mathcal{L}_{sty} = \sum_j||G_j^\phi(\hat{I}_t) - G_j^\phi(I_t)  ||_1
\end{equation}
where $G_j^\phi$ denotes the Gram matrix calculated from $\phi_j$. Finally, we use a standard adversarial loss $\mathcal{L}_{adv}$:
\begin{equation}
    \mathcal{L}_{adv} = \mathbb{E}[log(1-D(G(\{I_s^k\},\{P_s^k\}, P_t)))] + \mathbb{E}[log(D(I_t))] 
\end{equation}

% We also propose a fuse loss $\mathcal{L}_{fuse}$ to constrain the residual flow learning. Let $v_s^k$ and $v_t$ denotes the features generated by a specific layer of VGG19. Let $v_{s,w}^k$ be the feature warped by the refined flow. The fuse loss calculate the $L_{1}$ distance of the fused source VGG features and the target VGG feature:
% \begin{equation}
%     \mathcal{L}_{fuse} = \lambda_{fuse}(|\sum_{k=1}^{K}(v_{s,w}^k \cdot a^k) - v_t|_1)
% \end{equation}

\subsection{Implementation Details}
We implement our model using PyTorch~\cite{paszke2019pytorch} framework on a PC with four NVIDIA GTX 2080Ti GPUs. We adopt the Adam optimizer($\beta_1=0.9, \beta_2=0.999$) with a learning rate of 0.0001. The batch size is fixed to 5 for all tasks except Market-1501, in which batch size is set to 8. For the network details, please refer to the supplementary material.

\begin{figure*}[tb]
\includegraphics[width=\textwidth]{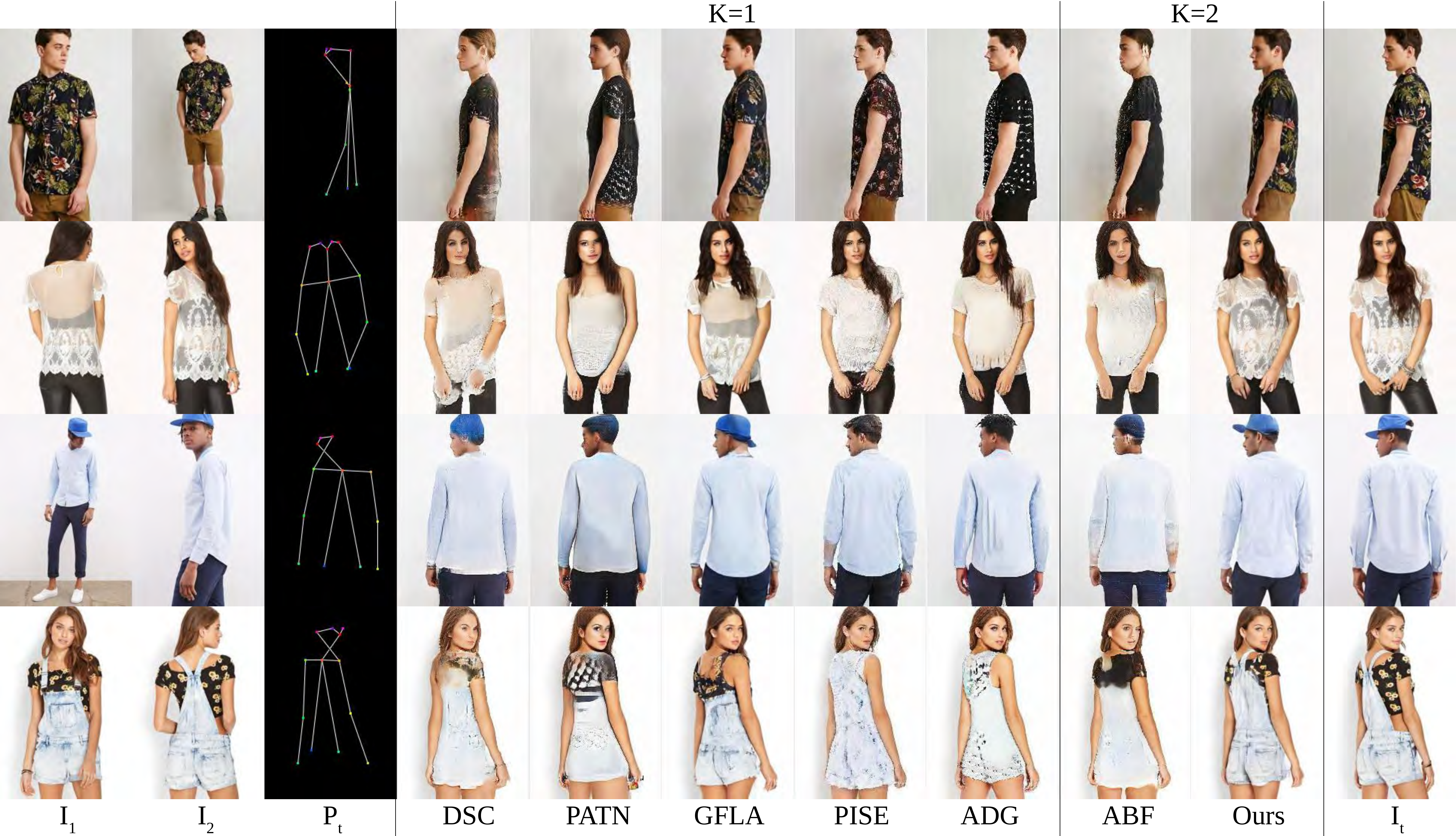}
\caption{Qualitative comparison on the DeepFashion dataset. DSC~\cite{Siarohin_2018_CVPR}, PATN~\cite{zhu2019progressive}, GFLA~\cite{ren2020deep}, PISE~\cite{zhang2021pise}, ADG~\cite{men2020controllable} are single source based methods, in which only I$_1$ is used as input. ABF~\cite{lathuiliere2020attention} and Ours are multi-source based methods, in which I$_1$ and I$_2$ are used as inputs. The last column shows the ground truth image.}
\label{fig_compare}
\end{figure*}
\section{Experiments}
\subsubsection{Datasets.} Since our model is designed to be general, we conduct experiments on three different tasks including pose transfer, view synthesis, and facial expression transfer on five challenging datasets. For human pose transfer, we use DeepFashion In-shop Clothes Retrieval Benchmark~\cite{liuLQWTcvpr16DeepFashion} and person re-identification dataset Market-1501~\cite{zheng2015scalable}. For novel view synthesis, we use real-world scenes (KITTI Visual Odometry Dataset~\cite{Geiger2012CVPR}) and rendered objects (ShapeNet chair dataset~\cite{chang2015shapenet}). For facial expression transfer, we use the talking videos dataset Voxceleb2\cite{Chung18b}. More details on the dataset can be found in the supplementary material.

These tasks are challenging in different ways. Human pose transfer needs to handle deformable human bodies with full and partial views, along with details on clothing; view synthesis needs to consider complex image semantics and features such as shadows; facial expression transfer needs to model consistent and realistic facial features. To handle them under one method, they show the generality of our model.
\subsubsection{Metrics.} How to evaluate the generated images remains an open problem in generative models. We follow~\cite{ren2020deep, zhang2021pise} and calculate the Frechet Inception Distance (FID)~\cite{heusel2017gans} and Learned Perceptual Image Patch Similarity(LPIPS)\cite{zhang2018unreasonable} to evaluate the performance of our model. For Market-1501, we further report the Mask-LPIPS (MLPIPS) score proposed in \cite{NIPS2017_34ed066d} to exclude the influence of the background. Besides, we perform a user study to evaluate the visual quality of the generated images.

\subsection{Qualitative Results}
We first show our results on the DeepFashion Benchmark with two input images in Fig.~\ref{fig:teaser} Left. In all cases, the target contains a novel view and pose, which means the model has to learn to correctly align feature of different sources. Further, the clothing details are also transferred well (e.g. the shirts in row 1, the hat in row 2,3), thanks to our multi-level feature modeling. Fig.~\ref{fig:teaser} Right shows the results on facial expression transfer on Voxceleb2. Realistic unseen expressions are generated by our model and source identities are preserved (row 1). Face in new head poses can also be generated. More results on KITTI and ShapeNet can be found in the supplementary material.

% In Fig.\ref{fig:ours_shapenet}, we show our results on ShapeNet chairs and KITTI urban images. For KITTI, our method can deal with large transformations in row 4-5. The source images has an obvious view difference from the target, and our method can predict reasonable transformations to warp the sources into target (see the black car from far to near and the shadow on the road). Also, novel views can be generated (row 1-3).

% For the ShapeNet chairs, our method can generate new view angles with small (row 1,3-5) and large differences from sources (row 2,6). Thin structures (e.g. Legs of chairs) are kept well (row 4,5). 

\subsection{Comparisons}

We compare our method with a variety of baseline methods across all tasks. For the human pose transfer task, we compare our approach with Def-GAN~\cite{Siarohin_2018_CVPR}, PATN~\cite{zhu2019progressive}, GFLA~\cite{ren2020deep}, PISE~\cite{zhang2021pise}, ADG~\cite{men2020controllable}, ABF~\cite{lathuiliere2020attention}. All baselines except ABF are single source based methods. For ABF, we train and evaluate their model using the same train/test split. For the other baseline methods, we use the pre-trained models and evaluate the performance on the testing set directly. For the novel view synthesis task, we compare our method with M2N~\cite{sun2018multiview}. We run the pre-trained model offered by the author to get results on KITTI and ShapeNet chair dataset. For the face generation task, we compare our method with NH-FF(the feed forward result of ~\cite{Zakharov_2019_ICCV}). We implement their method and report results using the same train/test split of Voxceleb2. 

\begin{table}[tb]
\centering
\setlength{\tabcolsep}{1.5mm}{
\begin{tabular}{|c|c|c|c|c|c|c|}
\hline
&&\multicolumn{2}{c|}{DeepFashion} & \multicolumn{3}{c|}{Market-1501}\\
\hline
  K & Model & FID & LPIPS & FID & LPIPS & MLPIPS \\
\hline
\multirow{7}*{1}   &DSC & 21.542    & 0.2384    & 24.861 & 0.2984  & 0.1495 \\
   &PATN   & 20.632    & 0.2553    & 22.753  & 0.3181  & 0.1585  \\
   &GFLA  & \underline{9.872}     & \underline{0.1963}    & \underline{19.750}  & \underline{0.2817} & 0.1483 \\
   &ADG & 14.476    & 0.2253    & -  & -  & - \\
  &PISE  & 11.524    & 0.2077    & -  & -  & -\\
   &ABF &27.303 & 0.2753& 32.588& 0.3015 &\underline{0.1480}\\
  & Ours  &  \textbf{9.750}  & \textbf{0.1867} & \textbf{17.362}  & \textbf{0.2730}
  &\textbf{0.1418}  \\

\hline
\multirow{2}*{2}    &ABF & 23.529 & 0.2577 & 30.274 & 0.2878 & 0.1390\\
    &Ours & \textbf{10.135}    & \textbf{0.1766}    &\textbf{15.716}& \textbf{0.2668} & \textbf{0.1339}\\
\hline
\multirow{2}*{3}    &ABF &  26.759   &0.2376 & 33.270& 0.2870&0.1314 \\
    &Ours &  \textbf{12.785}   & \textbf{0.1689}    & \textbf{16.263} &\textbf{0.2604} & \textbf{0.1277}\\
\hline
\end{tabular}}
\caption{Quantitative comparison with SOTA methods on the DeepFashion dataset and the Market-1501 dataset.}
\label{table_pose}
\end{table}

For the human image generation task, our method outperforms all baseline methods on all the metrics on both single-source and multi-source settings by a large margin, shown in Table~\ref{table_pose}. The LPIPS scores drop significantly when more source images are used, which demonstrates the effectiveness of utilizing multiple sources. Our model also shows superiority on other datasets, as shown in Table~\ref{table_chair}, ~\ref{table_face}.

The qualitative results on the DeepFashion dataset are shown in Fig.~\ref{fig_compare}. The baseline methods~\cite{zhu2019progressive, Siarohin_2018_CVPR, men2020controllable, zhang2021pise, lathuiliere2020attention} fail to keep the source appearance when the clothes pattern is complex (e.g. they fail to capture the texture details of the clothes in row 1). Flow-based method~\cite{ren2020deep} can preserve the details of the source, but struggles when there is a big gap between the source pose and the target one(e.g. It fails to generate correct back view from the front view in row 2, and predicts a wrong hat direction in row 3). ~\cite{lathuiliere2020attention} utilizes multiple inputs to generate the target view but fails to maintain source details in the generated images. In contrast, our model transfers high-fidelity details from the source images, and extracts information from different sources to overcome the single source ambiguity.

% \begin{figure}[tb]
% \includegraphics[width=1.0\linewidth]{kitti}
% \caption{Qualitative comparisons with \cite{sun2018multiview} on KITTI dataset.}
% \label{fig_compare_kitti}
% \end{figure}

\begin{table}[tb]
\centering
\begin{tabular}{|c|c|c|c|c|c|}
\hline
&&\multicolumn{2}{c|}{Chair}& \multicolumn{2}{c|}{KITTI}\\
\hline
  K & Model & FID & LPIPS & FID & LPIPS \\
\hline  
\multirow{2}*{2} &M2N &28.876 &0.1155 &18.798& 0.1958\\
 &Ours &\textbf{10.123} &\textbf{0.09607} &\textbf{8.505}&\textbf{0.1721} \\
\hline  
\multirow{2}*{4} &M2N &21.920 & 0.0901 & -&-\\
 &Ours &\textbf{7.697} &\textbf{0.0729} & -&- \\
\hline  
\end{tabular}
\caption{Quantitative comparison with M2N~\cite{sun2018multiview} on the KITTI dataset and the ShapeNet chair dataset. Our method gets lower FID scores and LPIPS scores than M2N on both datasets.}
\label{table_chair}
\end{table}
\begin{table}[tb]
\centering
\begin{tabular}{|c|c|c|c|}
\hline
  K & Model & FID & LPIPS  \\
\hline  
\multirow{2}*{2} &NH-FF &37.266 & 0.3150\\
 &Ours &\textbf{7.100} &\textbf{0.2130} \\
\hline
\multirow{2}*{4} &NH-FF &37.457 & 0.3131\\
 &Ours &\textbf{7.690} &\textbf{0.2084} \\
\hline  
\end{tabular}
\caption{Quantitative comparison with NH-FF~\cite{Zakharov_2019_ICCV} on the Voxceleb2 dataset. Our method outperforms NH-FF in both metrics.}
\label{table_face}
\end{table}

Qualitative comparisons on other datasets (Voxceleb2, KITTI and ShapeNet chair) can be found in the supplementary material.

\subsection{Ablation Study}
We present an ablation study on the human pose transfer task to clarify the impact of each part of our proposed method. 

\textbf{Baseline.} 
Our baseline model is U-Net architecture with feature warping, with no residual flow, attention map or occlusion map. Source features are fused by averaging.

\textbf{Without occlusion (w/o. occ).} 
This model is designed to see if the occlusion maps can benefits the learning. We remove the occlusion maps and source features are aggregated by attention maps.

\textbf{Without attention (w/o. attn).} 
The model is designed to see if the attention maps are effective in modeling confidence. We replace the attention maps of different views with the same value. The occlusion mechanism is adopted.

\textbf{Without residual flow (w/o. res).} 
In this configuration, we remove the residual block in the decoder to evaluate its contribution. Attention and occlusion mechanisms are employed in this model. 

\begin{figure}[t]
\includegraphics[width=1.0\linewidth]{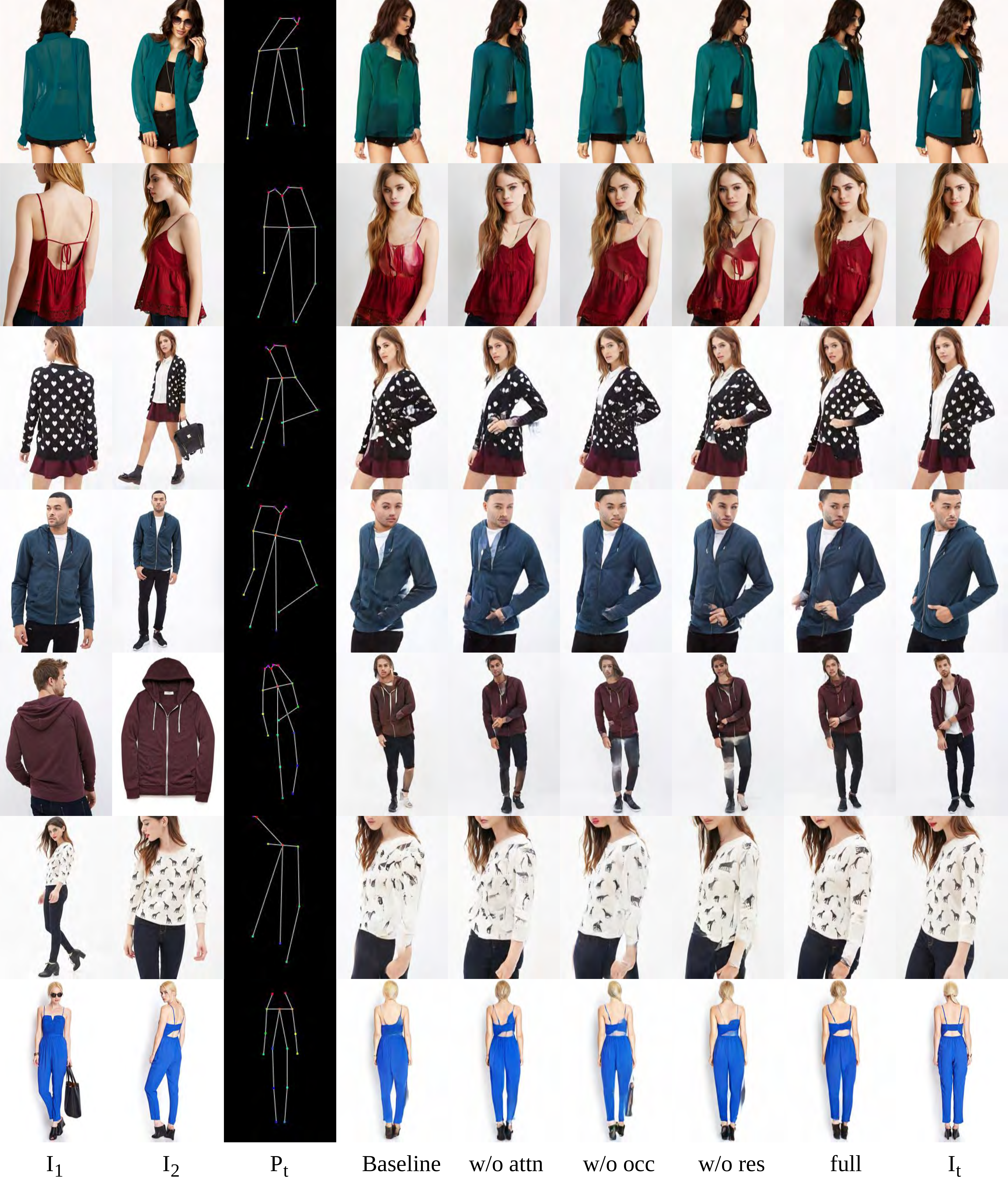}
\caption{Ablation study. The first two columns show the inputs, the third column shows the target pose and the last column shows the target image. Results of different ablation models are shown in the middle. Zoom in for better details.}
\label{fig_ablation}
\end{figure}

Qualitative results are shown in Fig.~\ref{fig_ablation}. The baseline method struggles when the source images have large differences in poses and views (e.g. the inner clothes in row 1, the dress in row 2), as it simply performs a weighted average over source features, without considering the relevant importance of each source. The baseline model also suffers from generating occluded parts and wrongly warped areas (e.g. face/arms of the man in row 5, hands of the woman in row 6). By adding the occlusion mechanism, the model can get improvements in these areas, but the model without attention mechanism tends to generate ghosting effects when the two sources are similar but at different scales (e.g. the collar of the man in row 4).  The model with attention maps but without occlusion maps could also fail when the model synthesizes new contents (e.g. the  the hand of the model in row 3, the face of the model in row 5,). And for the model without residual flow, the synthesized results suffer from blurry textures(the white spot in row 3, the vest in row 7) and irregular boundary of cloths(the boundary of the sweater in row 6). Detailed quantitative results are shown in Table~\ref{table:ablation}.

\begin{table}[tb]
\centering
\setlength{\tabcolsep}{1.3mm}{
\begin{tabular}{|c|c|c|c|c|c|c|}
\hline
&&\multicolumn{2}{c|}{DeepFashion} & \multicolumn{3}{c|}{Market-1501}\\
\hline
  K & Model & FID & LPIPS & FID & LPIPS & MLPIPS \\
\hline
\multirow{5}*{2}    &BaseLine & 11.078 & 0.1857 & 15.298 & 0.2714 & 0.1393\\
    &w/o attn & 10.835 & 0.1774 & 15.820 & 0.2676 & 0.1368\\
    &w/o occ & 10.421 & 0.1830 & \textbf{15.249} & 0.2689 & 0.1349\\
    &w/o res & 10.292 & 0.1777 & 17.701 & 0.2679 & 0.1483\\
    &full & \textbf{10.135}    & \textbf{0.1766}    &15.716& \textbf{0.2668} & \textbf{0.1339}\\
\hline
\multirow{2}*{3}    &w/o res &  13.306   &0.1710 & 18.092& 0.2645&0.1302 \\
    &full &  \textbf{12.785}   & \textbf{0.1679}    & \textbf{16.263} &\textbf{0.2604} &\textbf{0.1277}\\
    \hline

\end{tabular}}
\caption{Ablation Study on the DeepFashion dataset and Market-1501 dataset. Our full model achieves better performance than the baselines on both datasets. }
\label{table:ablation}
\end{table}

% The quantitative results can be found in Table~\ref{table:ablation}. The model with occlusion maps and target encoder gets significant improvements. The target pose encoder passes target pose information directly to the decoder, and the occlusion maps can capture the occluded areas and indicate the the wrongly warped region to be inpainted by the network. In addition, improvements are obtained via the attention mechanism, which allows the model to focus on the correctly warped regions and combine different sources by a relative weight. The residual fusing block gives the network the ability to further optimizing each source branch. Comparing to the model without the residual fusing block, the full model gets further performance gain when the number of input increases (e.g. The FID and LPIPS scores of the model w/o residual block get limited improvements when the number of inputs increase from 3 to 10 on the Market-1501 dataset).

\subsection{User Study}
We also conduct a user study to assess the visual quality. For each dataset, 30 volunteers are asked to accomplish two tasks: the first is a 'real or generated' test, following the protocol in ~\cite{NIPS2017_34ed066d, Siarohin_2018_CVPR}. For each model, volunteers are shown 55 real and 55 generated images in a random order. The volunteers are asked to judge whether the displayed image is real or generated in one second. The other is a comparison task. The volunteers are asked to finish 55 questions on each dataset, each question containing image pairs generated by ours method and a baseline method, with the same source images and target pose. The volunteers are asked to choose the one with better quality. All samples are randomly selected. Overall, our method significantly outperforms the baseline methods. Detailed results are shown in the supplementary material.

% \begin{table}[tb]
% \centering
% \begin{tabular}{|c|c|c|c|c|c|}
% \hline
%   K &Task  & Model & G2R & R2G & Judged as better  \\
% \hline  
% \multirow{2}*{2} &\multirow{2}*{Fashion} &ABF & 9.97 &22.64 & 10.85\\
% & &Ours& 49.48 &23.80 & \textbf{89.15}\\
% \hline
% \multirow{2}*{5} & \multirow{2}*{Market} &ABF& 35.26 &30.41 & 31.79\\
% & &Ours & 47.38 &30.91 & \textbf{68.21}\\
% \hline  
% \multirow{2}*{4}& \multirow{2}*{Voxceleb2}&NeuralHead  & 68.32 &31.07 & 32.01\\
% & & Ours& 74.54 &24.46 & \textbf{67.99}\\
% \hline  
% \multirow{2}*{4}& \multirow{2}*{ShapeNet Chair}&Multi2novel  & 43.80 &44.68 & 28.60\\
% & &Ours& 47.55 &45.79 & \textbf{71.40}\\
% \hline  
% \multirow{2}*{2} & \multirow{2}*{KITTI}&Multi2novel & 42.48 & 20.00 & 16.42\\
% & &Ours & 69.48 &24.46 & \textbf{83.58}\\
% \hline  
% \end{tabular}
% \caption{User study (\%) results on different tasks.}
% \label{table_user_study}
% \end{table}

\section{Conclusion}
We have proposed a new general method for multi-source image generation. Given a guiding pose, our framework effectively rectifies the issues caused by the misalignment among the sources, which makes it widely applicable to datasets with in-the-wild images taken by un-calibrated cameras. The model generality has been tested on a variety of vastly different datasets including human poses, street scenes, faces and 3D objects, and verified by its universal successes. In exhaustive comparisons, our model outperforms the state-of-the-art methods in various tasks.

\section{Acknowledgements}
We thank the reviewers for their comments and suggestions for improving the paper. The work was supported by NSF China (No. 61772462, No. U1736217, No. 61772457), the 100 Talents Program of Zhejiang University, and NSF (2016414 and 2011471).
% \nobibliography{aaai22}
\bibliography{aaai22}

\begin{thebibliography}{30}
\providecommand{\natexlab}[1]{#1}

\bibitem[{Balakrishnan et~al.(2018)Balakrishnan, Zhao, Dalca, Durand, and
  Guttag}]{balakrishnan2018synthesizing}
Balakrishnan, G.; Zhao, A.; Dalca, A.~V.; Durand, F.; and Guttag, J. 2018.
\newblock Synthesizing images of humans in unseen poses.
\newblock In \emph{Proceedings of the IEEE Conference on Computer Vision and
  Pattern Recognition}, 8340--8348.

\bibitem[{Brock, Donahue, and Simonyan(2018)}]{brock2018large}
Brock, A.; Donahue, J.; and Simonyan, K. 2018.
\newblock Large scale GAN training for high fidelity natural image synthesis.
\newblock \emph{arXiv preprint arXiv:1809.11096}.

\bibitem[{Bulat and Tzimiropoulos(2017)}]{bulat2017far}
Bulat, A.; and Tzimiropoulos, G. 2017.
\newblock How far are we from solving the 2D \& 3D Face Alignment problem? (and
  a dataset of 230,000 3D facial landmarks).
\newblock In \emph{International Conference on Computer Vision}.

\bibitem[{Chan et~al.(2019)Chan, Ginosar, Zhou, and Efros}]{Chan_2019_ICCV}
Chan, C.; Ginosar, S.; Zhou, T.; and Efros, A.~A. 2019.
\newblock Everybody Dance Now.
\newblock In \emph{Proceedings of the IEEE/CVF International Conference on
  Computer Vision (ICCV)}.

\bibitem[{Chang et~al.(2015)Chang, Funkhouser, Guibas, Hanrahan, Huang, Li,
  Savarese, Savva, Song, Su et~al.}]{chang2015shapenet}
Chang, A.~X.; Funkhouser, T.; Guibas, L.; Hanrahan, P.; Huang, Q.; Li, Z.;
  Savarese, S.; Savva, M.; Song, S.; Su, H.; et~al. 2015.
\newblock Shapenet: An information-rich 3d model repository.
\newblock \emph{arXiv preprint arXiv:1512.03012}.

\bibitem[{Chung, Nagrani, and Zisserman(2018)}]{Chung18b}
Chung, J.~S.; Nagrani, A.; and Zisserman, A. 2018.
\newblock VoxCeleb2: Deep Speaker Recognition.
\newblock In \emph{INTERSPEECH}.

\bibitem[{Geiger, Lenz, and Urtasun(2012)}]{Geiger2012CVPR}
Geiger, A.; Lenz, P.; and Urtasun, R. 2012.
\newblock Are we ready for Autonomous Driving? The KITTI Vision Benchmark
  Suite.
\newblock In \emph{Conference on Computer Vision and Pattern Recognition
  (CVPR)}.

\bibitem[{Han et~al.(2019)Han, Hu, Huang, and Scott}]{Han_2019_ICCV}
Han, X.; Hu, X.; Huang, W.; and Scott, M.~R. 2019.
\newblock ClothFlow: A Flow-Based Model for Clothed Person Generation.
\newblock In \emph{Proceedings of the IEEE/CVF International Conference on
  Computer Vision (ICCV)}.

\bibitem[{Heusel et~al.(2017)Heusel, Ramsauer, Unterthiner, Nessler, and
  Hochreiter}]{heusel2017gans}
Heusel, M.; Ramsauer, H.; Unterthiner, T.; Nessler, B.; and Hochreiter, S.
  2017.
\newblock Gans trained by a two time-scale update rule converge to a local nash
  equilibrium.
\newblock \emph{Advances in neural information processing systems}, 30.

\bibitem[{Johnson, Alahi, and Fei-Fei(2016)}]{Johnson2016Perceptual}
Johnson, J.; Alahi, A.; and Fei-Fei, L. 2016.
\newblock Perceptual losses for real-time style transfer and super-resolution.
\newblock In \emph{European Conference on Computer Vision}.

\bibitem[{Lathuili{\`e}re et~al.(2020)Lathuili{\`e}re, Sangineto, Siarohin, and
  Sebe}]{lathuiliere2020attention}
Lathuili{\`e}re, S.; Sangineto, E.; Siarohin, A.; and Sebe, N. 2020.
\newblock Attention-based fusion for multi-source human image generation.
\newblock In \emph{Proceedings of the IEEE/CVF Winter Conference on
  Applications of Computer Vision}, 439--448.

\bibitem[{Liu et~al.(2019)Liu, Xu, Zollhoefer, Kim, Bernard, Habermann, Wang,
  and Theobalt}]{liu2019neural}
Liu, L.; Xu, W.; Zollhoefer, M.; Kim, H.; Bernard, F.; Habermann, M.; Wang, W.;
  and Theobalt, C. 2019.
\newblock Neural rendering and reenactment of human actor videos.
\newblock \emph{ACM Transactions on Graphics (TOG)}, 38(5): 1--14.

\bibitem[{Liu et~al.(2016)Liu, Luo, Qiu, Wang, and
  Tang}]{liuLQWTcvpr16DeepFashion}
Liu, Z.; Luo, P.; Qiu, S.; Wang, X.; and Tang, X. 2016.
\newblock DeepFashion: Powering Robust Clothes Recognition and Retrieval with
  Rich Annotations.
\newblock In \emph{Proceedings of IEEE Conference on Computer Vision and
  Pattern Recognition (CVPR)}.

\bibitem[{Ma et~al.(2017)Ma, Jia, Sun, Schiele, Tuytelaars, and
  Van~Gool}]{NIPS2017_34ed066d}
Ma, L.; Jia, X.; Sun, Q.; Schiele, B.; Tuytelaars, T.; and Van~Gool, L. 2017.
\newblock Pose Guided Person Image Generation.
\newblock In Guyon, I.; Luxburg, U.~V.; Bengio, S.; Wallach, H.; Fergus, R.;
  Vishwanathan, S.; and Garnett, R., eds., \emph{Advances in Neural Information
  Processing Systems}, volume~30. Curran Associates, Inc.

\bibitem[{Men et~al.(2020)Men, Mao, Jiang, Ma, and Lian}]{men2020controllable}
Men, Y.; Mao, Y.; Jiang, Y.; Ma, W.-Y.; and Lian, Z. 2020.
\newblock Controllable Person Image Synthesis with Attribute-Decomposed GAN.
\newblock In \emph{Computer Vision and Pattern Recognition (CVPR), 2020 IEEE
  Conference on}.

\bibitem[{Paszke et~al.(2019)Paszke, Gross, Massa, Lerer, Bradbury, Chanan,
  Killeen, Lin, Gimelshein, Antiga et~al.}]{paszke2019pytorch}
Paszke, A.; Gross, S.; Massa, F.; Lerer, A.; Bradbury, J.; Chanan, G.; Killeen,
  T.; Lin, Z.; Gimelshein, N.; Antiga, L.; et~al. 2019.
\newblock Pytorch: An imperative style, high-performance deep learning library.
\newblock \emph{Advances in neural information processing systems}, 32:
  8026--8037.

\bibitem[{Ren et~al.(2020)Ren, Yu, Chen, Li, and Li}]{ren2020deep}
Ren, Y.; Yu, X.; Chen, J.; Li, T.~H.; and Li, G. 2020.
\newblock Deep image spatial transformation for person image generation.
\newblock In \emph{Proceedings of the IEEE/CVF Conference on Computer Vision
  and Pattern Recognition}, 7690--7699.

\bibitem[{Ren et~al.(2019)Ren, Yu, Zhang, Li, Liu, and
  Li}]{ren2019structureflow}
Ren, Y.; Yu, X.; Zhang, R.; Li, T.~H.; Liu, S.; and Li, G. 2019.
\newblock Structureflow: Image inpainting via structure-aware appearance flow.
\newblock In \emph{Proceedings of the IEEE/CVF International Conference on
  Computer Vision}, 181--190.

\bibitem[{Siarohin et~al.(2018)Siarohin, Sangineto, Lathuilière, and
  Sebe}]{Siarohin_2018_CVPR}
Siarohin, A.; Sangineto, E.; Lathuilière, S.; and Sebe, N. 2018.
\newblock Deformable GANs for Pose-Based Human Image Generation.
\newblock In \emph{The IEEE Conference on Computer Vision and Pattern
  Recognition (CVPR)}.

\bibitem[{Sun et~al.(2018)Sun, Huh, Liao, Zhang, and Lim}]{sun2018multiview}
Sun, S.-H.; Huh, M.; Liao, Y.-H.; Zhang, N.; and Lim, J.~J. 2018.
\newblock Multi-view to Novel View: Synthesizing Novel Views with Self-Learned
  Confidence.
\newblock In \emph{European Conference on Computer Vision}.

\bibitem[{Ulyanov, Vedaldi, and Lempitsky(2016)}]{ulyanov2016instance}
Ulyanov, D.; Vedaldi, A.; and Lempitsky, V. 2016.
\newblock Instance normalization: The missing ingredient for fast stylization.
\newblock \emph{arXiv preprint arXiv:1607.08022}.

\bibitem[{Wang et~al.(2019)Wang, Liu, Tao, Liu, Kautz, and
  Catanzaro}]{wang2018fewshotvid2vid}
Wang, T.-C.; Liu, M.-Y.; Tao, A.; Liu, G.; Kautz, J.; and Catanzaro, B. 2019.
\newblock Few-shot Video-to-Video Synthesis.
\newblock In \emph{Advances in Neural Information Processing Systems
  (NeurIPS)}.

\bibitem[{Wang et~al.(2018)Wang, Liu, Zhu, Liu, Tao, Kautz, and
  Catanzaro}]{wang2018vid2vid}
Wang, T.-C.; Liu, M.-Y.; Zhu, J.-Y.; Liu, G.; Tao, A.; Kautz, J.; and
  Catanzaro, B. 2018.
\newblock Video-to-Video Synthesis.
\newblock In \emph{Conference on Neural Information Processing Systems
  (NeurIPS)}.

\bibitem[{Xingjian et~al.(2015)Xingjian, Chen, Wang, Yeung, Wong, and
  Woo}]{xingjian2015convolutional}
Xingjian, S.; Chen, Z.; Wang, H.; Yeung, D.-Y.; Wong, W.-K.; and Woo, W.-c.
  2015.
\newblock Convolutional LSTM network: A machine learning approach for
  precipitation nowcasting.
\newblock In \emph{Advances in neural information processing systems},
  802--810.

\bibitem[{Zakharov et~al.(2019)Zakharov, Shysheya, Burkov, and
  Lempitsky}]{Zakharov_2019_ICCV}
Zakharov, E.; Shysheya, A.; Burkov, E.; and Lempitsky, V. 2019.
\newblock Few-Shot Adversarial Learning of Realistic Neural Talking Head
  Models.
\newblock In \emph{Proceedings of the IEEE/CVF International Conference on
  Computer Vision (ICCV)}.

\bibitem[{Zhang et~al.(2021)Zhang, Li, Lai, and Yang}]{zhang2021pise}
Zhang, J.; Li, K.; Lai, Y.-K.; and Yang, J. 2021.
\newblock PISE: Person Image Synthesis and Editing with Decoupled GAN.
\newblock \emph{arXiv preprint arXiv:2103.04023}.

\bibitem[{Zhang et~al.(2018)Zhang, Isola, Efros, Shechtman, and
  Wang}]{zhang2018unreasonable}
Zhang, R.; Isola, P.; Efros, A.~A.; Shechtman, E.; and Wang, O. 2018.
\newblock The unreasonable effectiveness of deep features as a perceptual
  metric.
\newblock In \emph{Proceedings of the IEEE conference on computer vision and
  pattern recognition}, 586--595.

\bibitem[{Zheng et~al.(2015)Zheng, Shen, Tian, Wang, Wang, and
  Tian}]{zheng2015scalable}
Zheng, L.; Shen, L.; Tian, L.; Wang, S.; Wang, J.; and Tian, Q. 2015.
\newblock Scalable Person Re-identification: A Benchmark.
\newblock In \emph{Computer Vision, IEEE International Conference on}.

\bibitem[{Zhou et~al.(2016)Zhou, Tulsiani, Sun, Malik, and
  Efros}]{zhou2016view}
Zhou, T.; Tulsiani, S.; Sun, W.; Malik, J.; and Efros, A.~A. 2016.
\newblock View Synthesis by Appearance Flow.
\newblock In \emph{European Conference on Computer Vision}.

\bibitem[{Zhu et~al.(2019)Zhu, Huang, Shi, Yu, Wang, and
  Bai}]{zhu2019progressive}
Zhu, Z.; Huang, T.; Shi, B.; Yu, M.; Wang, B.; and Bai, X. 2019.
\newblock Progressive pose attention transfer for person image generation.
\newblock In \emph{Proceedings of the IEEE/CVF Conference on Computer Vision
  and Pattern Recognition}, 2347--2356.

\end{thebibliography}

% \appendix
\clearpage

% \appendixpage
% \title{Pose Guided Image Generation from Misaligned Sources via Residual Flow Based
% Correction\\
% Supplementary Material}

% \author{
%     %Authors
%     % All authors must be in the same font size and format.
%     Anonymous AAAI submission\\
%     Paper ID 4099
% }

\section{Implementation Details}
% We first introduce some notations of basic network components: $IN$(instance normalization), $BN$(batch normalization), $conv\_k\_s$(convolutional layer with kernel size $k$ and stride $s$).  

Our network includes three parts: source appearance extractor, flow confidence extractor and target image predictor. Their detailed architecture is shown in  Table~\ref{tab:image_feature_extractor},~\ref{tab:flow_extrator},~\ref{tab:target_generator} respectively, where 
\begin{itemize}
    \item \textbf{ResBlockD} denotes a down-sampling residual block following~\cite{brock2018large} where The ReLU activation is replaced by Leaky-ReLU.
    \item \textbf{ResBlockU} denotes an up-sampling residual block following~\cite{brock2018large} where the ReLU activation is replaced by Leaky-ReLU, and batch normalization layers are replaced by instance normalization~\cite{ulyanov2016instance}.
    \item \textbf{ResBlock} is a residual block keeping the spatial resolution.
    \item \textbf{ResFusingBlock} is the proposed residual fusing module. The only learnable part is the residual flow predictor in the residual module, which is a single convolutional layer. We use two ResFusingBlocks in all experiments.
    \item \textbf{Skip} is a skip connection adding the feature maps of an encoding layer and decoding layer with the same spatial resolution.
\end{itemize}
We train our model in two stages. The flow confidence extractor is first warmed up to generate flow fields. Then we train the whole network is in an end-to-end manner. We set $\lambda_{per}=0.25$, $\lambda_{sty}=250$, $\lambda_{l_1}=2.5$, $\lambda_{cor}=2.5$, and $\lambda_{reg}=0.001$ across all the experiments.

\section{Datasets Details}
We conduct experiments on three different tasks including pose transfer, view synthesis, and facial expression transfer on five challenging datasets: 
\begin{itemize}
    \item \textbf{Human pose transfer}. We experiment on two datasets: DeepFashion In-shop Clothes Retrieval Benchmark~\cite{liuLQWTcvpr16DeepFashion} and person re-identification dataset Market-1501~\cite{zheng2015scalable}. DeepFashion contains 52712 number of in-shop model images with various poses and clothes, with a resolution of $256\times256$ pixels.  Rather than generating all the possible pairs for all the identities, we generate tuples of size  $K+1$ ($K$ source images and $1$ target image) in the same way as
    ~\cite{lathuiliere2020attention} for fair comparison. The identities of the training and the testing sets do not overlap. 
    Market-1501 contains 32688 low resolution ($128\times64$) images of 1,501 identities. Images of each identity are captured by at most six cameras, with tremendous pose/background/illumination variety. The aforementioned method is applied to create training tuples. 
    
    \item \textbf{Novel view synthesis}. We experiment on two datasets in this task: Real-world scenes (KITTI Visual Odometry Dataset~\cite{Geiger2012CVPR}) and rendered objects (ShapeNet chair dataset~\cite{chang2015shapenet}). KITTI contains frame sequences with a resolution of $256\times256$ captured by a car traveling through city scenes with camera poses, the camera pose is a 6-DOF vector containing translation and rotation of the camera. ShapeNet Chair is composed of render images of chair models with a dimension of $256\times256$ in 54 viewpoints , which corresponds to 18 azimuth angles (sampled in range $[0,340]$ with 20-degree increments) and the elevations of 0, 10, and 20. The pose of each image is represented as a concatenation of two one-hot vectors: an 18 element vector indicating the azimuth angle and a 3 element vector indicating the elevation. For fair comparison, We follow~\cite{sun2018multiview} and perform training and testing with $K=2$ on the KITTI dataset and $K=4$ for ShapeNet chair dataset. 
    \item \textbf{Facial expression transfer}. We use a large talking head videos dataset Voxceleb2\cite{Chung18b} in this task, which has 5,994 identities for training set and 118 for testing. The video sequences are 224p at 25 fps. We adopt the off-the-shelf face alignment code~\cite{bulat2017far} to crop the frames and obtain the landmarks as in~\cite{Zakharov_2019_ICCV}. Then, we sample 8 images for each video from the original training set to get our multi-source image training set.

\end{itemize}

\begin{table}[t]
    \centering
    \renewcommand\arraystretch{1.5}
    \begin{tabular}{|c|}
           \hline
           
           Source Image $I_s^k$ $(3 \times {H}\times{W})$\\\hline

           ResBlockD $(3 \times {H}\times{W}) \rightarrow (64 \times \frac{H}{2}\times\frac{W}{2})$ \\\hline

           $^\ast$ ResBlockD $(64 \times \frac{H}{2}\times\frac{W}{2})\rightarrow (128 \times \frac{H}{4}\times\frac{W}{4})$  \\\hline

           $^\ast$ ResBlockD $(128 \times \frac{H}{4}\times\frac{W}{4})\rightarrow(256 \times \frac{H}{8}\times\frac{W}{8})$ \\\hline
           
    \end{tabular}
    \caption{Architecture of the Source Appearance Extractor, $\ast$ indicates the output layer.}
    \label{tab:image_feature_extractor}
\end{table}  

\begin{table}[t]
    \centering
    \renewcommand\arraystretch{1.5}
    \setlength{\tabcolsep}{1.3mm}{
    \begin{tabular}{|c|}
           \hline

           Source Image $I_s^k$ $(3 \times {H}\times{W})$\\
           Source Pose $P_s^k$ $(c_p \times {H}\times{W})$\\
           Target Pose $P_t$ $(c_p \times {H}\times{W})$\\\hline

           ResBlockD $((3+c_p\times2)\times {H}\times{W}) \rightarrow$ $(32\times \frac{H}{2}\times\frac{W}{2})$ \\\hline

           ResBlockD $(32\times \frac{H}{2}\times\frac{W}{2}) \rightarrow(64\times \frac{H}{4}\times\frac{W}{4})$ \\\hline

           ResBlockD $(64\times \frac{H}{4}\times\frac{W}{4}) \rightarrow(128\times \frac{H}{8}\times\frac{W}{8})$\\\hline

           ResBlockD $(128\times \frac{H}{8}\times\frac{W}{8}) \rightarrow(256\times \frac{H}{16}\times\frac{W}{16})$ \\\hline

           ResBlockD $(256\times \frac{H}{16}\times\frac{W}{16}) \rightarrow(512\times \frac{H}{32}\times\frac{W}{32})$ \\\hline

           (ResBlockU $512\times \frac{H}{32}\times\frac{W}{32} \rightarrow 256\times \frac{H}{16}\times\frac{W}{16}$) + Skip 
           \\\hline

           (ResBlockU $256\times \frac{H}{16}\times\frac{W}{16} \rightarrow 128\times \frac{H}{8}\times\frac{W}{8}$) + Skip \\
           $^\ast$ Conv $(128\times \frac{H}{8}\times\frac{W}{8}) \rightarrow (2\times\frac{H}{8}\times\frac{W}{8})$  \\
           $^\ast$ Conv $(128\times \frac{H}{8}\times\frac{W}{8}) \rightarrow (1\times\frac{H}{8}\times\frac{W}{8})$  \\
           $^\ast$ Conv $(128\times \frac{H}{8}\times\frac{W}{8}) \rightarrow (1\times\frac{H}{8}\times\frac{W}{8})$  \\\hline

           (ResBlockU $128\times \frac{H}{16}\times\frac{W}{16} \rightarrow 64\times \frac{H}{4}\times\frac{W}{4}$) + Skip \\
           $^\ast$ Conv $(64\times \frac{H}{4}\times\frac{W}{4}) \rightarrow (2\times\frac{H}{4}\times\frac{W}{4})$  \\
           $^\ast$ Conv $(64\times \frac{H}{4}\times\frac{W}{4}) \rightarrow (1\times\frac{H}{4}\times\frac{W}{4})$  \\
           $^\ast$ Conv $(64\times \frac{H}{4}\times\frac{W}{4}) \rightarrow (1\times\frac{H}{4}\times\frac{W}{4})$  \\\hline

    \end{tabular}}
    \caption{Architecture of the Flow Confidence Extractor. $c_p$ denotes the number of channels for pose. $c_p=3$ for Voxceleb2 where pose is encoded as RGB landmark image. $c_p=18$ for DeepFashion and Market-1501 where pose is represented by 18-channel heatmap.  For ShapeNet and KITTI, the channel of pose is 21 and 6 respectively. $\ast$ indicates the output layer. At each spatial resolution, flow map, attention map as well as occlusion map are output to the Target Image Predictor.}
    \label{tab:flow_extrator}
\end{table}

\begin{table}[t]
    \centering
    \renewcommand\arraystretch{1.5}
    \begin{tabular}{|c|}
           \hline
           
           Target Pose $P_t$ ($c_p \times {H}\times{W}$)\\\hline

           ResBlockD $(c_p \times {H}\times{W})$ $\rightarrow$ $(64\times \frac{H}{2}\times\frac{W}{2})$ \\\hline

           ResBlockD $(64\times \frac{H}{2}\times\frac{W}{2})$ $\rightarrow$ ($128\times \frac{H}{4}\times\frac{W}{4}$) \\\hline

           ResBlockD $(128\times \frac{H}{4}\times\frac{W}{4})$ $\rightarrow$ $(256\times \frac{H}{8}\times\frac{W}{8})$ \\\hline
           
           ResFusingBlock $(256\times \frac{H}{8}\times\frac{W}{8})$ $\rightarrow$ $(256\times \frac{H}{8}\times\frac{W}{8})$ \\
           ResBlock $(256\times \frac{H}{8}\times\frac{W}{8})$ $\rightarrow$ $(256\times \frac{H}{8}\times\frac{W}{8})$\\
           ResBlockU $(256\times \frac{H}{8}\times\frac{W}{8})$ $\rightarrow$ $(128\times \frac{H}{4}\times\frac{W}{4})$\\\hline
           
           ResFusingBlock $(128\times \frac{H}{4}\times\frac{W}{4})$ $\rightarrow$ $(128\times \frac{H}{4}\times\frac{W}{4})$ \\
           ResBlock $(128\times \frac{H}{4}\times\frac{W}{4})$ $\rightarrow$ $(128\times \frac{H}{4}\times\frac{W}{4})$\\
           ResBlockU $(128\times \frac{H}{4}\times\frac{W}{4})$ $\rightarrow$ $(64\times \frac{H}{2}\times\frac{W}{2})$\\\hline

           ResBlock $(64\times \frac{H}{2}\times\frac{W}{2})$ $\rightarrow$ $(64\times \frac{H}{2}\times\frac{W}{2})$\\
           ResBlockU $(64\times \frac{H}{2}\times\frac{W}{2})$ $\rightarrow$ $(64\times {H}\times{W})$\\\hline

           Conv $(64\times {H}\times{W})$ $\rightarrow$ $(3\times {H}\times{W})$\\
           $^\ast$ Tanh\\\hline
           
    \end{tabular}
    \caption{Architecture of the target image predictor. $c_p$ denotes the number of channels for pose. \textbf{ResFusingBlock} takes the output from \textbf{Flow confidence extractor} as well as \textbf{Source appearance extractor} with the same spatial resolution and output the fused feature. $\ast$ indicates the output layer.}
    \label{tab:target_generator}
\end{table}

\section{More results}
\begin{figure*}
\centering
    \includegraphics[width=1.0\linewidth]{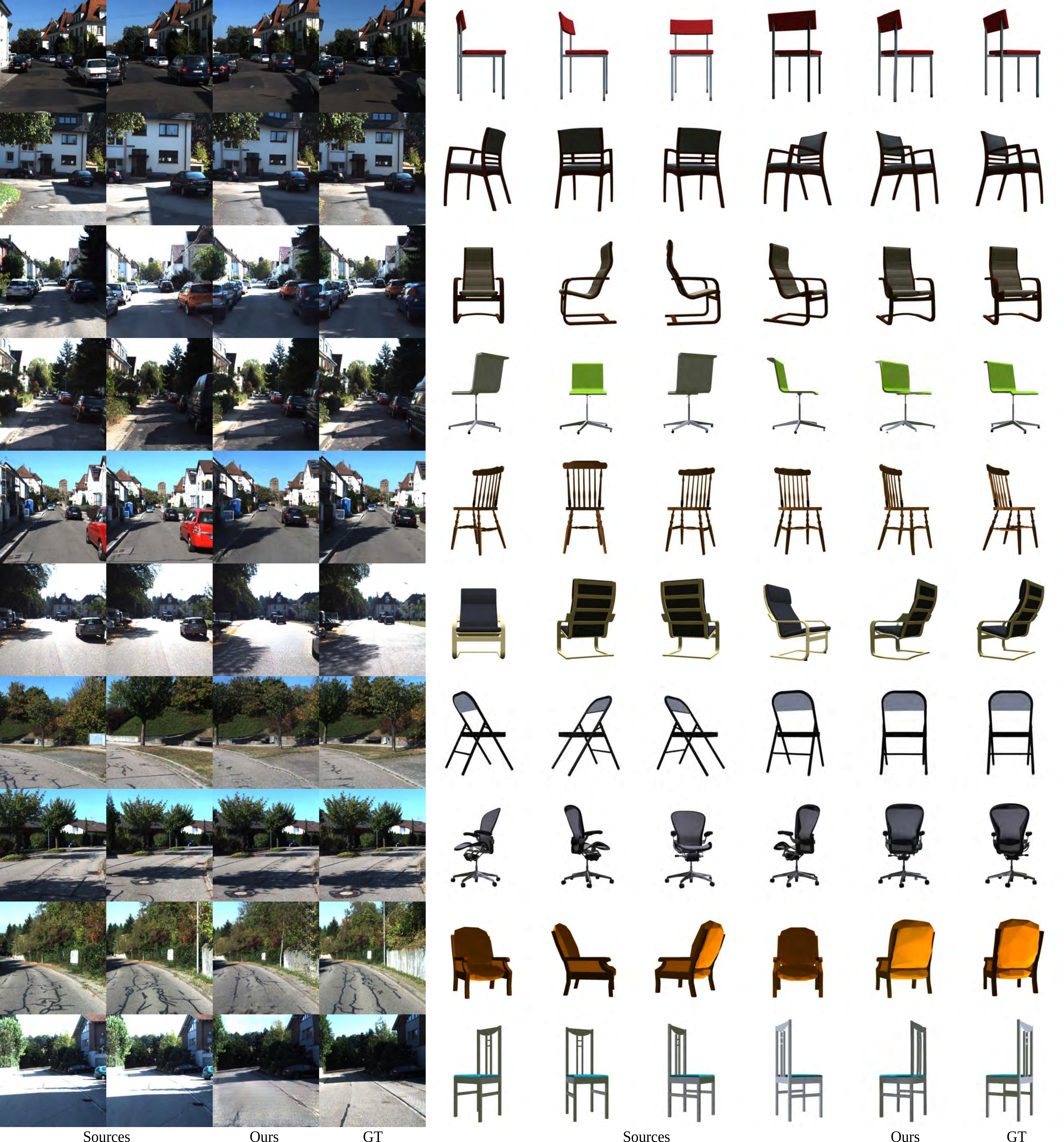}
    \caption{Qualitative results on KITTI dataset and ShapeNet chair dataste. The left part shows the results on KITTI, and the right part shows the results on ShapeNet chairs. The first two/four columns shows the input images, the last two columns shows our results and the ground truth images. }
    \label{fig:ours_shapenet}
\end{figure*}

\noindent\textbf{Results of novel view synthesis.}
In Fig.\ref{fig:ours_shapenet}, we show our results on novel view synthesis on two datasets. As shown on the left of the figure, although the source images have significant view differences from the target, our method can predict reasonable transformations to warp the sources into target (see the black car from far to near and the shadow on the road). Also, novel views can be generated (the last row).
For the ShapeNet chairs, our method can generate new view angles with small (row 1,3-5) and large differences from sources (row 2,6). Thin structures (e.g. Legs of chairs) are kept well (row 4,5). 

% Fig.~\ref{fig:ours_face} shows the results on facial expression transfer, on the Voxceleb2 dataset, unseen expressions are well generated by our model and source identities are preserved. New head poses/views can also be generated (row 6, 7, 9, 10). Adding more source images can considerably improve the quality and identity of the result (row 1,3-7,9-11).

\begin{figure*}
    \centering
        \includegraphics[width=0.98\linewidth]{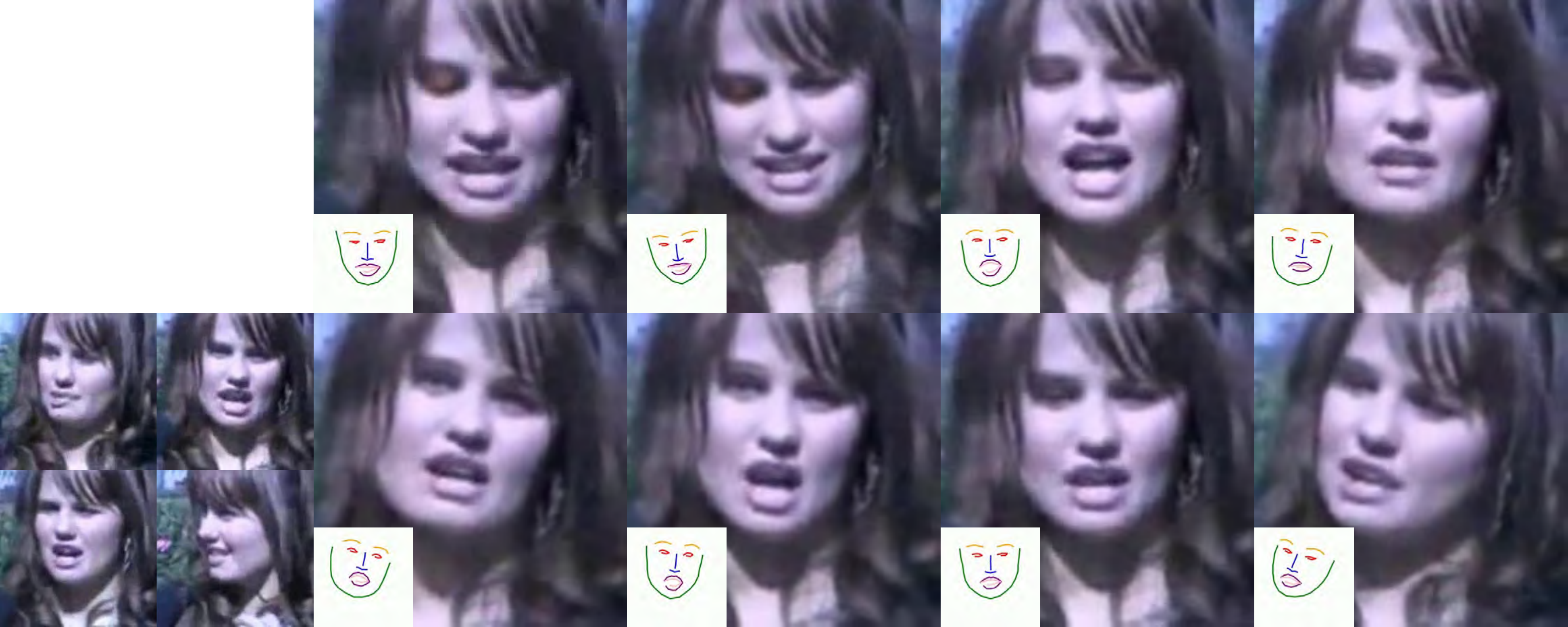}
        \caption{Visualization of face sequence generation. The input 4 images are shown on the bottom left. The 8 images on the right are generated by the corresponding facial landmarks. Zoom in for better details.\LJW{Youtube} }
        \label{fig:face_video}
    \end{figure*}

\begin{figure*}
    \centering
        \includegraphics[width=0.98\linewidth]{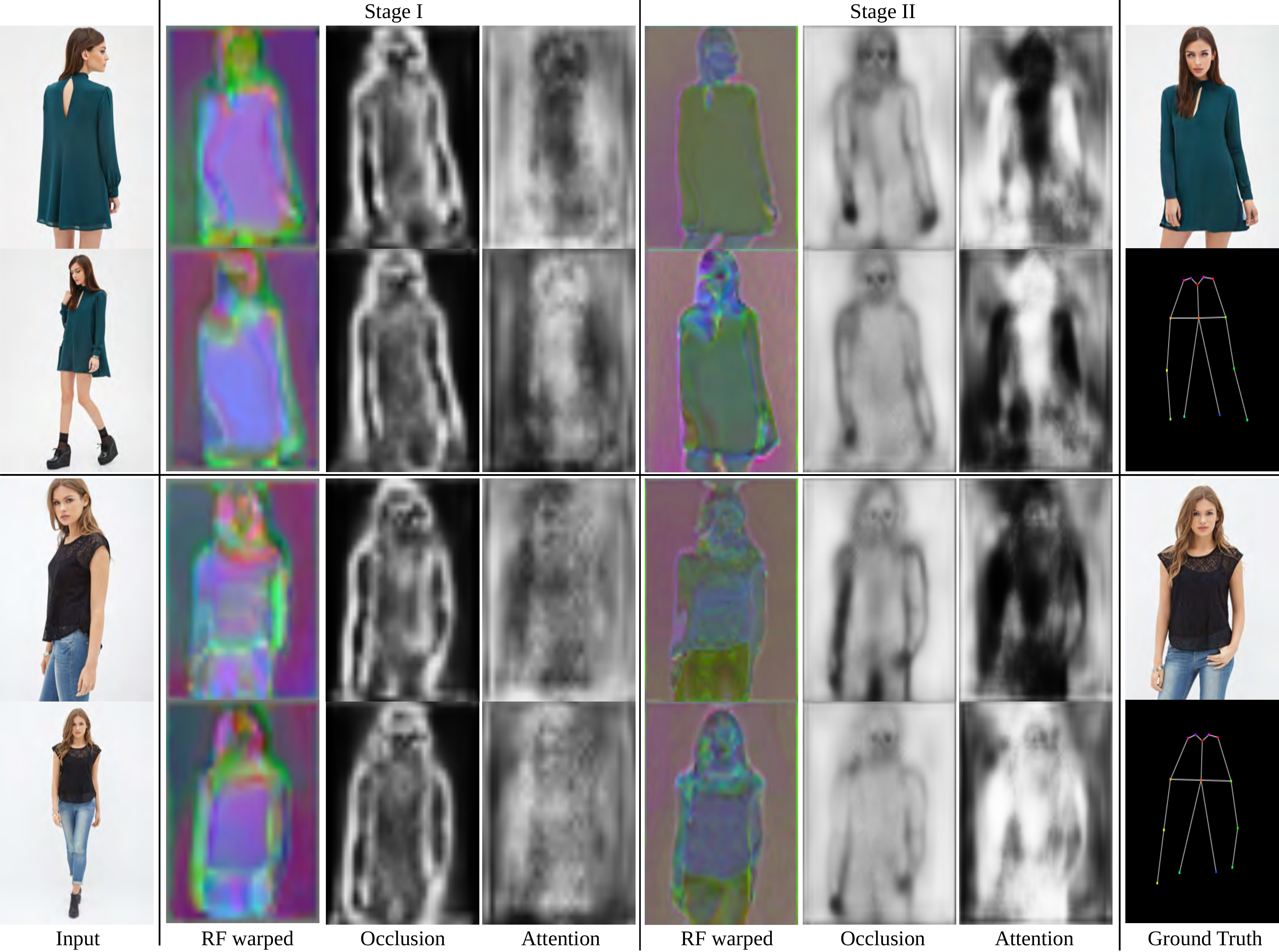}
        \caption{Visualization of warped features, occlusion, and attention in the RF modules at different stages. We up-sample the features to the image resolution for visualization. Zoom in for better details. }
        \label{fig:fig_feature}
    \end{figure*}

\noindent\textbf{Results on face videos.}
We have conducted some experiments using talking videos from VoxCeleb2, and the results are shown in Fig.~\ref{fig:face_video}. In general, our model could generate consistent appearance in various poses given multiple sources, which suggests its potential on videos if the model is trained with more temporal constraints and other prior knowledge. Video is available at this link \footnote{https://youtu.be/Zv3RMxFdVLU}. \\
\noindent\textbf{Feature visualization.}
We visualize the intermediate features (occlusion, attention, flows) in Fig.~\ref{fig:fig_feature}.

\section{More comparisons on different datasets}

\begin{figure*}
\centering
\includegraphics[width=0.9\linewidth]{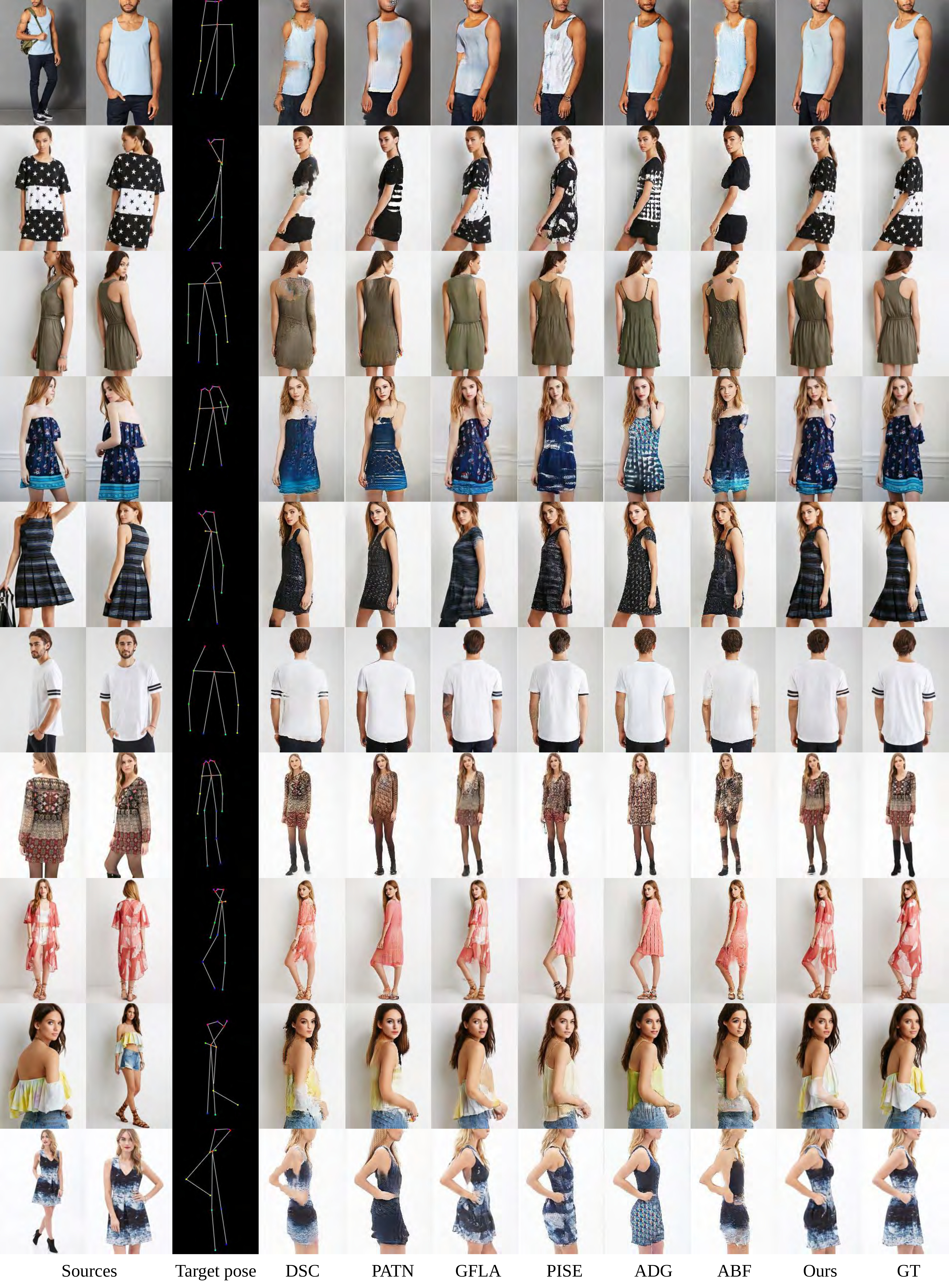}
\caption{Comparisons with SOTA methods on the DeepFashion dataset. DSC\cite{Siarohin_2018_CVPR}, PATN\cite{zhu2019progressive}, GFLA\cite{ren2020deep}, PISE\cite{zhang2021pise}, ADG\cite{men2020controllable} are single source based methods, in which only $I_1$ is used as input. ABF and Ours are multi-source based methods, in which $I_1$ and $I_2$ are used as inputs. The last column shows the ground truth image. Zoom in for better details.}
\label{fig_compare_supp}
\end{figure*}

In the human image generation task, our method outperforms all the other state-of-the-art methods on all the metrics on both single source and multi-source settings by a large margin, shown in the Table~\ref{table_pose_supp}. The LPIPS scores drop significantly while we use more input source images, which demonstrates the effectiveness of utilizing multiple sources. More qualitative results on DeepFashion dataset are shown in Figure~\ref{fig_compare_supp}. 

\begin{figure*}
\centering
\includegraphics[width=0.85\linewidth]{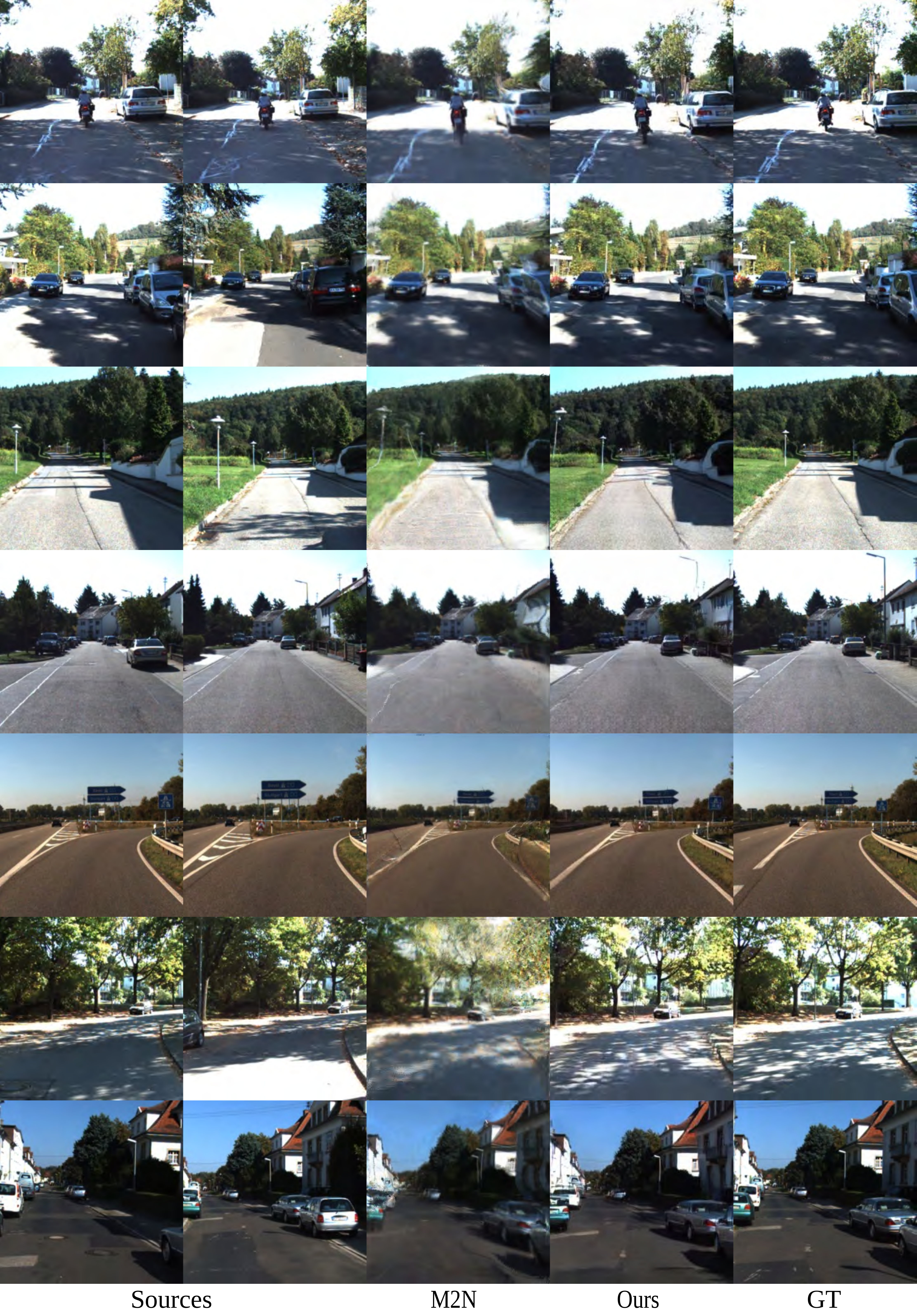}
\caption{Qualitative comparisons with \cite{sun2018multiview} on KITTI dataset.}
\label{fig_compare_kitti}
\end{figure*}

For the KITTI dataset, the comparisons are shown in Fig.\ref{fig_compare_kitti}. we generate sharper images than \cite{sun2018multiview} when the scenes go from far to near (row 2). In addition, their images have larger distortions as they predict dense flow maps at the resolution of the image (row 1-4). In contrast, we predict flows and residual flow maps at multiple feature levels, which leads to a better estimation of the flow field. Ghosting effects can be found in their results (row 3), where they generate ghosting road lights. We believe this is because we predict occlusion maps in each source and mask out the badly warped area with low confidence values.

\begin{figure*}
\centering
\includegraphics[width=0.93\linewidth]{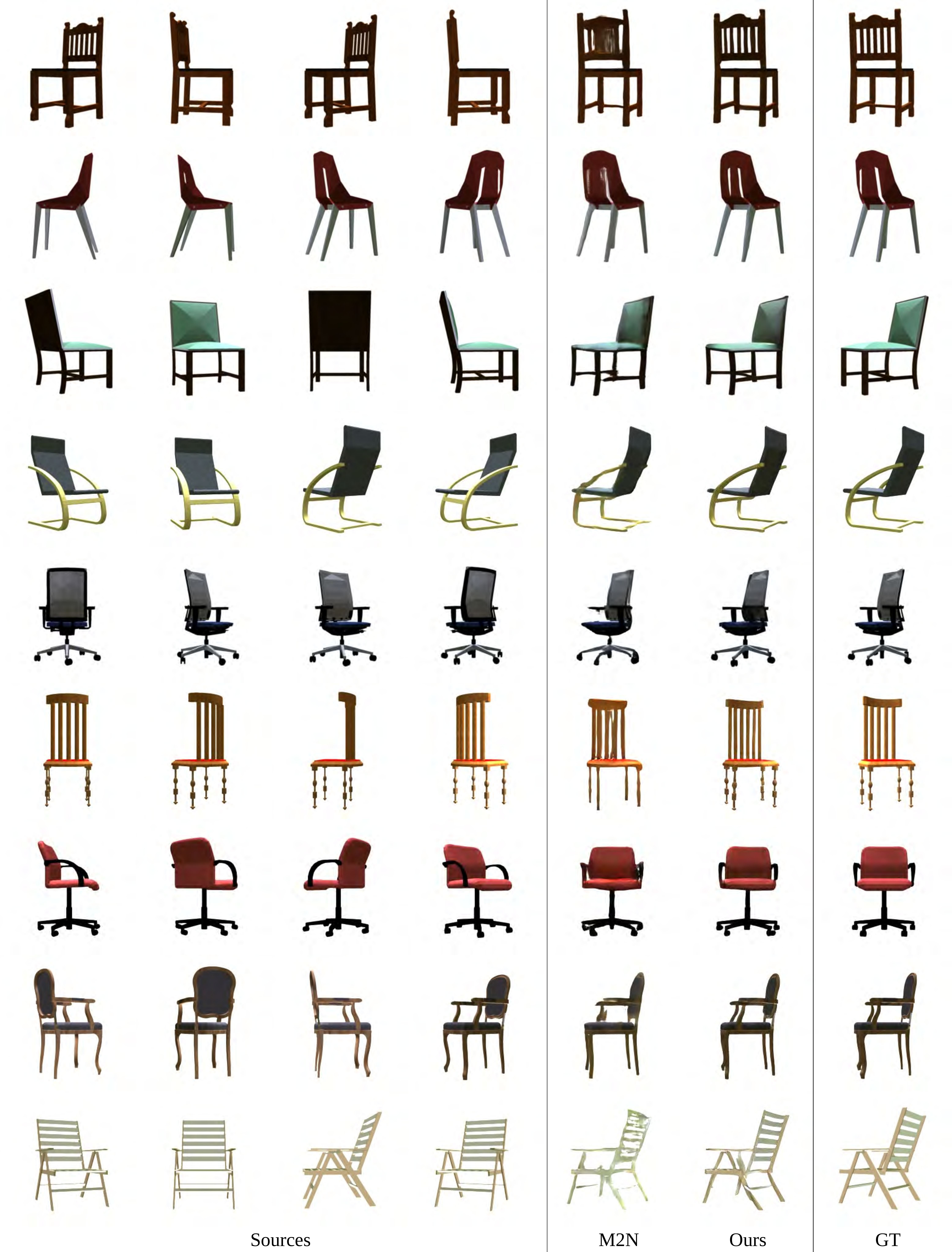}
\caption{Qualitative comparisons with M2N \cite{sun2018multiview} on ShapeNet chair dataset using 4 inputs}
\label{fig_compare_chair}
\end{figure*}

For the ShapeNet chairs, the comparison results are shown in Fig.~\ref{fig_compare_chair}. The thin structures generated by ours are preserved better than~\cite{sun2018multiview} (row 1-2,4-5). And ours results contain better details (e.g. the leg pattern in row 2).

\begin{figure*}
\centering
\includegraphics[width=1.0\linewidth]{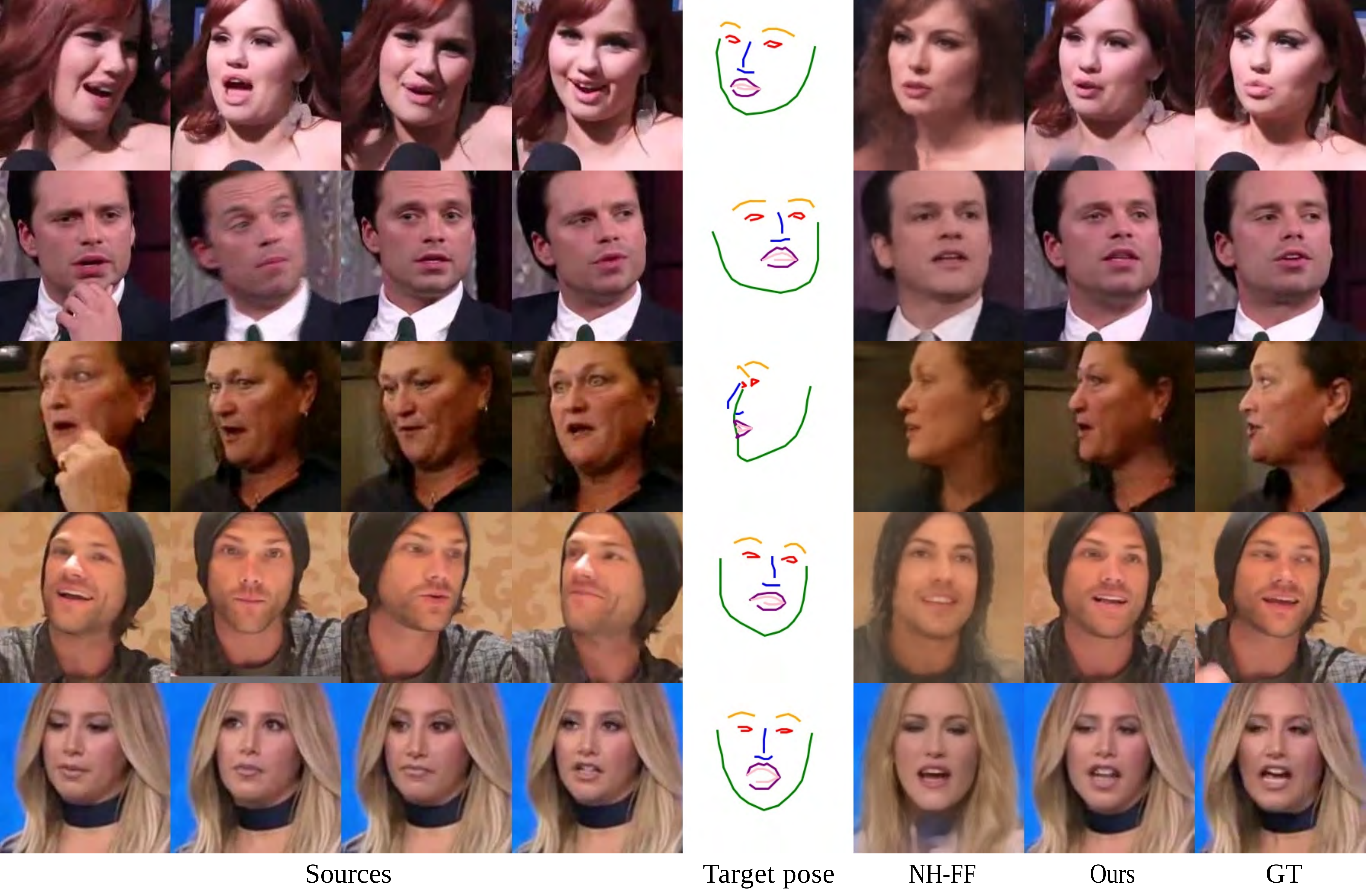}
\caption{Comparisons with NH-FF\cite{Zakharov_2019_ICCV} on Voxceleb2 Dataset}
\label{fig_compare_face}
\end{figure*}

For Voxceleb2, the comparison results are shown in Fig.~\ref{fig_compare_face}. Our method generates images with realistic details (e.g. the beard in row 2,4), while keeping the source identity. In contrast, the feed forward method~\cite{Zakharov_2019_ICCV} fails to keep the source identity, as they embed input images into a style vector and may lose the spatial information of the source images.

\begin{table}[hbt!]
\centering
\setlength{\tabcolsep}{1.3mm}{
\begin{tabular}{|c|c|c|c|c|c|c|}
\hline
&&\multicolumn{2}{c|}{DeepFashion} & \multicolumn{3}{c|}{Market-1501}\\
\hline
  K & Model & FID & LPIPS & FID & LPIPS & MLPIPS \\
\hline
\multirow{7}*{1}   &DSC & 21.542    & 0.2384    & 24.861 & 0.2984  & 0.1495 \\
   &PATN   & 20.632    & 0.2553    & 22.753  & 0.3181  & 0.1585  \\
   &GFLA  & \underline{9.872}     & \underline{0.1963}    & \underline{19.750}  & \underline{0.2817} & 0.1483 \\
   &ADG & 14.476    & 0.2253    & -  & -  & - \\
  &PISE  & 11.524    & 0.2077    & -  & -  & -\\
   &ABF &27.303 & 0.2753& 32.588& 0.3015 &\underline{0.1480}\\
  & Ours  &  \textbf{9.750}  & \textbf{0.1867} & \textbf{17.362}  & \textbf{0.2730}
  &\textbf{0.1418}  \\

\hline
\multirow{2}*{2}    &ABF & 23.529 & 0.2577 & 30.274 & 0.2878 & 0.1390\\
    &Ours & \textbf{10.135}    & \textbf{0.1766}    &\textbf{15.716}& \textbf{0.2668} & \textbf{0.1339}\\
\hline
\multirow{2}*{3}    &ABF &  26.759   &0.2376 & 33.270& 0.2870&0.1314 \\
    &Ours &  \textbf{12.785}   & \textbf{0.1689}    & \textbf{16.263} &\textbf{0.2604} & \textbf{0.1277}\\
\hline
\multirow{2}*{5}    &ABF &  -   &  -   &31.105 &0.2791 &0.1274\\
    &Ours &  -   &   -  &\textbf{15.169} &\textbf{0.2510} &\textbf{0.1186}\\
\hline
\multirow{2}*{7}    &ABF &   -  &  -   &51.940 &0.2953 &0.1244\\
    &Ours &  -   &  -   &\textbf{14.194} &\textbf{0.2423} &\textbf{0.1130}\\
\hline
\multirow{2}*{10}    &ABF &   -  &   -  &46.433 &0.2906 &0.1195\\
   &Ours &  -   &  -   &\textbf{14.242} &\textbf{0.2360} &\textbf{0.1075}\\
\hline
\end{tabular}}
\caption{Comparison with the state of the art on the DeepFashion dataset and Market-1501 dataset}
\label{table_pose_supp}
\end{table}

\section{Ablation study}
\begin{figure*}
\centering
    \includegraphics[width=1.0\linewidth]{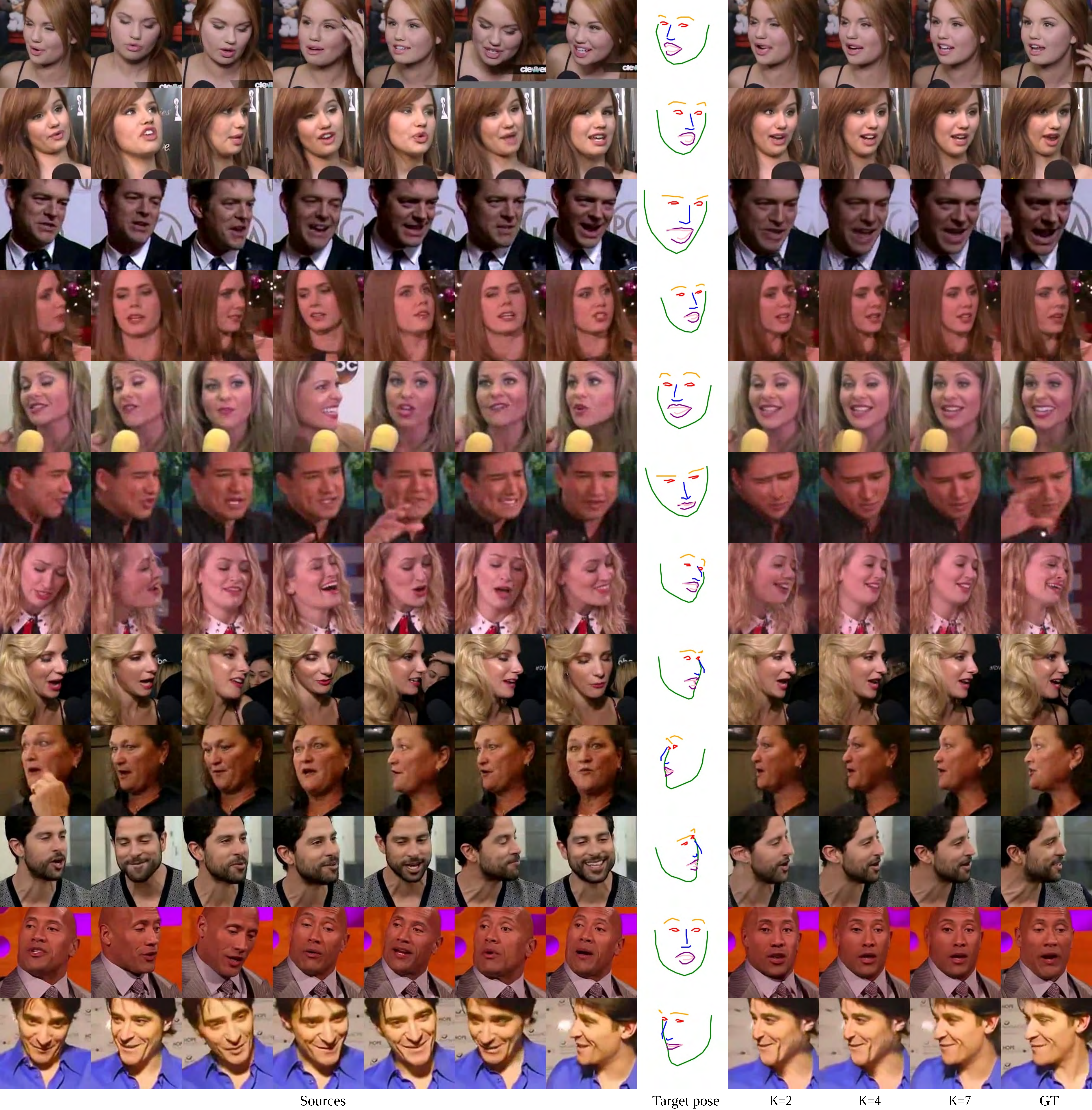}
    \caption{Qualitative results on Voxceleb2 dataset. The first seven columns shows the input images, Column 8 shows the target landmark, and Column 9-11 show the image generated by our method using the first two, four, seven images as input, respectively. The last column shows the ground truth. Zoom in for better details.}
    \label{fig:ours_face}
\end{figure*}

\begin{figure*}
\centering
    \includegraphics[width=0.7\linewidth]{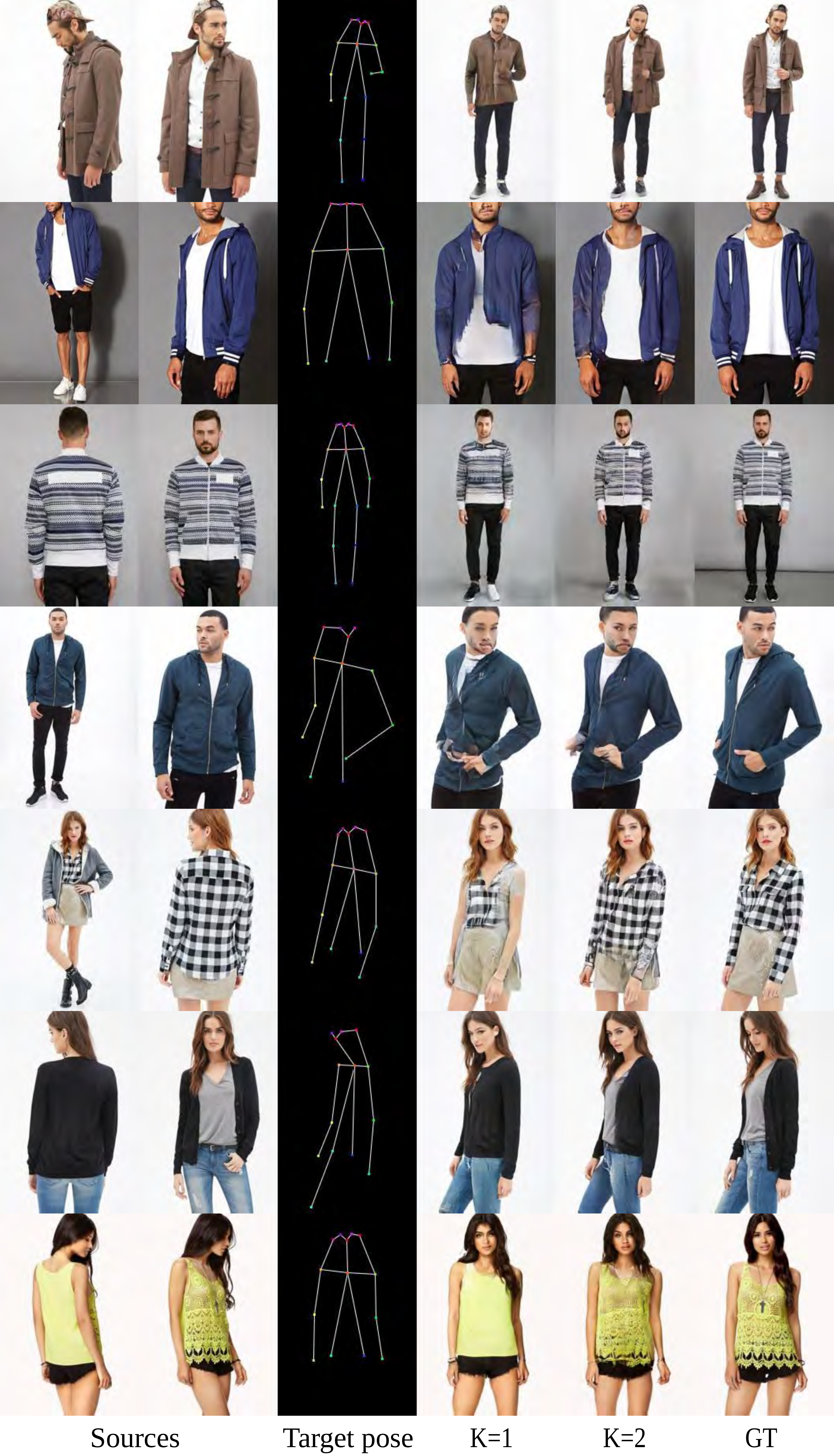}
    \caption{Qualitative results on DeepFashion dataset. The first two columns shows the input images, Column 3 shows the target landmark, and Column 4-5 show the image generated by our method using the first 1,2 images as input, respectively. The last column shows the ground truth. Zoom in for better details.}
    \label{fig:fashion}
\end{figure*}

More quantitative results are shown in Table~\ref{table_ablation_supp}
For qualitative results, we show the impact of different number of inputs on the generated images on Voxceleb2 dataset and DeepFashion dataset. As in Fig.~\ref{fig:ours_face}, we show the result of face image generation with $K=2,4,7$.In Fig.~\ref{fig:fashion}, we show the result of pose transfer with $K=1,2$ on DeepFashion.  Feeding more sources to the model considerably improves the visual quality.

\begin{table}[hbt!]
\centering
\setlength{\tabcolsep}{1.2mm}{
\begin{tabular}{|c|c|c|c|c|c|c|}
\hline
&&\multicolumn{2}{c|}{DeepFashion} & \multicolumn{3}{c|}{Market-1501}\\
\hline
  K & Model & FID & LPIPS & FID & LPIPS & MLPIPS \\
\hline
\multirow{5}*{2}    &Baseline & 11.078 & 0.1857 & \textbf{15.298} & 0.2714 & 0.1393\\
    &w/o attn & 10.835 & 0.1774 & 15.820 & 0.2676 & 0.1368\\
    &w/o occ & 10.421 & 0.1830 & - & - & -\\
    &w/o res & 10.292 & 0.1777 & 17.701 & 0.2679 & 0.1483\\
    &full & \textbf{10.135}    & \textbf{0.1766}    &15.716& \textbf{0.2668} & \textbf{0.1339}\\
\hline
\multirow{2}*{3}    &w/o res &  13.306   &0.1710 & 18.092& 0.2645&0.1302 \\
    &full &  \textbf{12.785}   & \textbf{0.1679}    & \textbf{16.263} &\textbf{0.2604} & \textbf{0.1277}\\
\hline
\multirow{2}*{5}    &w/o res &  -   &  -   &17.709 &0.2629 &0.1290\\
    &full &  -   &   -  &\textbf{15.169} &\textbf{0.2510} &\textbf{0.1186}\\
\hline
\multirow{2}*{7}    &w/o res &   -  &  -   &18.225 &0.2605 &0.1273\\
    &full &  -   &  -   &\textbf{14.194} &\textbf{0.2423} &\textbf{0.1130}\\
\hline
\multirow{2}*{10}    &w/o res &   -  &   -  &18.169 &0.2601 &0.1259\\
    &full &  -   &  -   &\textbf{14.242} &\textbf{0.2360} &\textbf{0.1075}\\
\hline
\end{tabular}}
\caption{Ablation study on more number of inputs}
\label{table_ablation_supp}
\end{table}
% As in Fig.~\ref{?}, we show the fashion image generated by different number of K. 

% \begin{figure}[tb]
%     \includegraphics[width=1.0\linewidth]{ours_fashion_2view}
%     \caption{Qualitative results on DeepFashion dataset. The first two columns shows the input images. Column 3 shows the target pose. Column 4 shows the results. The last column shows the ground truth image.}
%     \label{fig:ours_fashion}
% \end{figure}

% \section{More results}
% We show more results generated by our method on different dataset in this section.

% The baseline methods~\cite{zhu2019progressive, Siarohin_2018_CVPR, men2020controllable, zhang2021pise, lathuiliere2020attention} fail to keep the source appearance when the clothes pattern is complex (e.g. the texture of the clothes in row 1), because they do not use dense flow field to model the motion (\ref{fig_compare}). Flow-based method~\cite{ren2020deep} can preserve the details of the source, but they struggle when the target pose has large difference from the source pose (e.g. from the back view to generate the from view in row 4,6 and the hat direction in row 2). In contrast, our model generates images with high-fidelity details while keeping the source identity, and simultaneously deals with more inputs and extract useful information from different sources to compensate for occlusion. \cite{lathuiliere2020attention} utilize multiple inputs to generate the target view, however, they do not predict a dense flow field, which leads to blurry details in the generated images.

\section{User Study}
We show our user study results in Table~\ref{table_user_study}. Following \cite{NIPS2017_34ed066d}, for the 'real or generated' test, G2R denotes the rate of generated image being selected as real, R2G denotes the rate of real images selected as generated. 
For the comparison task, Preferred denotes which method obtains better visual quality. Our method significantly outperforms the baseline methods on these metrics.

\begin{table}[hbt!]
\centering
\begin{tabular}{|c|c|c|c|c|c|}
\hline
  K &Task  & Model & G2R & R2G & Preferred  \\
\hline  
\multirow{2}*{2} &\multirow{2}*{Fashion} &ABF & 9.97 &22.64 & 10.85\\
& &Ours& 49.48 &23.80 & \textbf{89.15}\\
\hline
\multirow{2}*{5} & \multirow{2}*{Market} &ABF& 35.26 &30.41 & 31.79\\
& &Ours & 47.38 &30.91 & \textbf{68.21}\\
\hline  
\multirow{2}*{4}& \multirow{2}*{Voxceleb2}&NH-FF  & 68.32 &31.07 & 32.01\\
& & Ours& 74.54 &24.46 & \textbf{67.99}\\
\hline  
\multirow{2}*{4}& \multirow{2}*{ShapeNet}&M2N  & 43.80 &44.68 & 28.60\\
& &Ours& 47.55 &45.79 & \textbf{71.40}\\
\hline  
\multirow{2}*{2} & \multirow{2}*{KITTI}&M2N & 42.48 & 20.00 & 16.42\\
& &Ours & 69.48 &24.46 & \textbf{83.58}\\
\hline  
\end{tabular}
\caption{User study (\%) results on different tasks.}
\label{table_user_study}
\end{table}

% \bibliography{aaai22}
\newpage
    

\end{document}